\pdfoutput=1
\documentclass[sigconf]{acmart}
\usepackage{graphicx}
\usepackage{caption}
\usepackage{subfig}
\usepackage{colortbl}

\AtBeginDocument{%
  }

\copyrightyear{2025}
\acmYear{2025}
\setcopyright{none}

\setcctype{by}
\acmConference[CHI '25]{CHI Conference on Human Factors in Computing Systems}{April 26-May 1, 2025}{Yokohama, Japan}
\acmBooktitle{CHI Conference on Human Factors in Computing Systems (CHI '25), April 26-May 1, 2025, Yokohama, Japan}\acmDOI{10.1145/3706598.3713779}
\acmISBN{979-8-4007-1394-1/25/04}

\begin{document}


\title{AEGIS: Human Attention-based Explainable Guidance for Intelligent Vehicle Systems}

\author{Zhuoli Zhuang}
\orcid{0009-0008-5088-3370}
\affiliation{%
  \institution{University of Technology Sydney}
  \city{Sydney}
  \state{NSW}
  \country{Australia}
}
\email{zhuoli.zhuang@student.uts.edu.au}

\author{Cheng-You Lu}
\affiliation{%
  \institution{University of Technology Sydney}
  \city{Sydney}
  \state{NSW}
  \country{Australia}}
\email{cheng-you.lu@student.uts.edu.au}

\author{Yu-Cheng Fred Chang}
\affiliation{%
  \institution{University of Technology Sydney}
  \city{Sydney}
  \state{NSW}
  \country{Australia}
}
\email{fred.chang@uts.edu.au}

\author{Yu-Kai Wang}
\affiliation{%
 \institution{University of Technology Sydney}
  \city{Sydney}
  \state{NSW}
  \country{Australia}}
  \email{yukai.wang@uts.edu.au}

\author{Thomas Do}
\affiliation{%
  \institution{University of Technology Sydney}
  \city{Sydney}
  \state{NSW}
  \country{Australia}}
  \email{thomas.do@uts.edu.au}

\author{Chin-Teng Lin}
\affiliation{%
  \institution{University of Technology Sydney}
  \city{Sydney}
  \state{NSW}
  \country{Australia}}
\email{chin-teng.lin@uts.edu.au}

\begin{abstract}

Improving decision-making capabilities in Autonomous Intelligent Vehicles (AIVs) has been a heated topic in recent years. 
Despite advancements, training machine to capture regions of interest for comprehensive scene understanding, like human perception and reasoning, remains a significant challenge. 
This study introduces a novel framework, Human \underline{A}ttention-based \underline{E}xplainable \underline{G}uidance for \underline{I}ntelligent Vehicle \underline{S}ystems (AEGIS\footnote{In Greek mythology, the Aegis is a protective shield associated with Zeus and Athena, symbolizing guidance and protection.}). 
AEGIS uses a pre-trained human attention model to guide reinforcement learning (RL) models to identify critical regions of interest for decision-making.
By collecting 1.2 million frames from 20 participants across six scenarios, AEGIS pre-trains a model to predict human attention patterns.
The learned human attention\footnote{We refer to the prediction of the human attention model as learned human attention.} guides the RL agent’s focus on task-relevant objects, prioritizes critical instances, enhances robustness in unseen environments, and leads to faster learning convergence.
This approach enhances interpretability by making machine attention more comparable to human attention and thus enhancing the RL agent’s performance in diverse driving scenarios. The code is available in \href{https://github.com/ALEX95GOGO/AEGIS}{https://github.com/ALEX95GOGO/AEGIS}.
\end{abstract}



\begin{teaserfigure}
  \includegraphics[width=\textwidth]{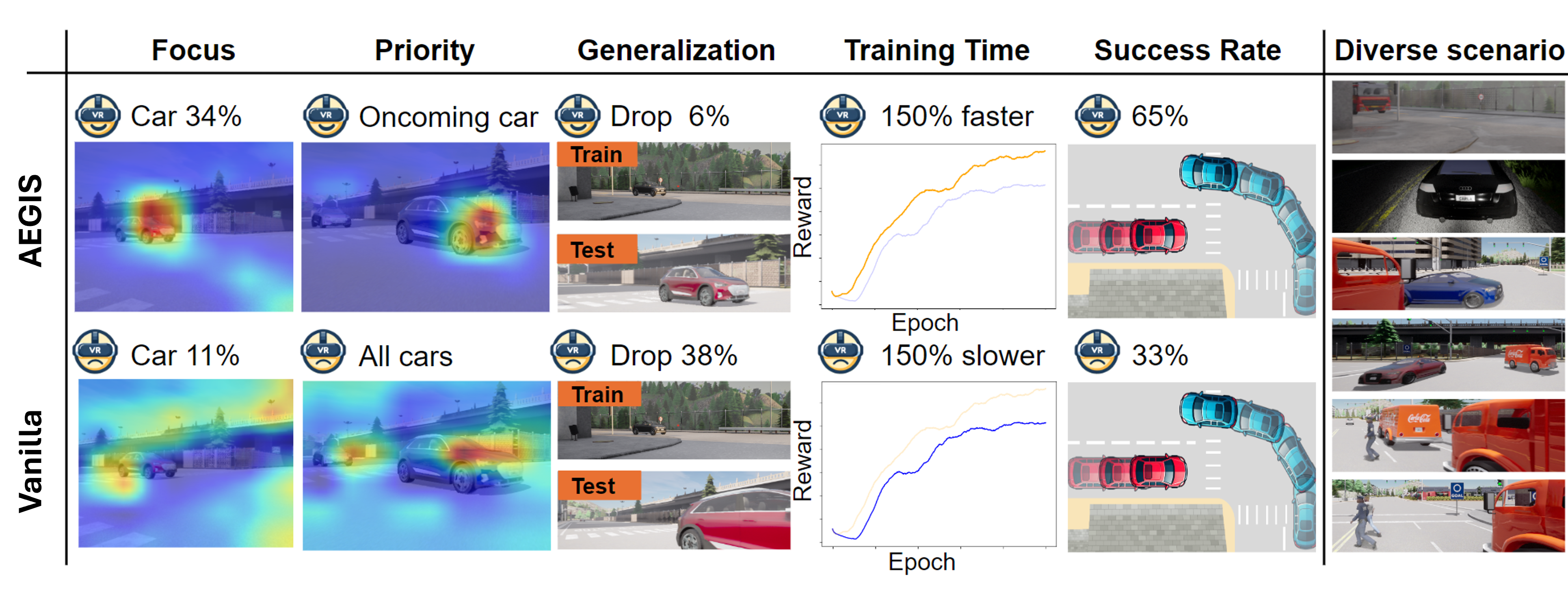}
  \caption{Comparison between AEGIS and RL without human attention guidance (Vanilla). AEGIS enhances focus on the regions of interest, prioritizes important objects, achieves robust performance in unseen environments, accelerates training speed, and achieves better performance for collision avoidance in the left-turn scenario. We analyze and benchmark our approach in six challenging scenarios: car-following, left-turn, and four diverse occlusion scenes.}
  \label{fig:teaser}
\end{teaserfigure}


\maketitle
\section{Introduction}
Deep learning methods for autonomous intelligent vehicles (AIVs) have been rapidly developing over the past two decades.
However, these models raise concerns about safety of AIVs due to the nontransparency of the decision-making process \cite{kuutti2020survey}.
Despite recent attempts to improve the transparency of deep learning models for AIVs \cite{shao2023reasonnet, shao2023safety}, these deep learning models lack explicit scene understanding and reasoning, the two basic cognitive skills of human drivers. 
Specifically, the decision-making of human drivers starts with visual perception via eye movements. 
In this way, task-relevant visual information can be extracted to better understand the scene. 
Moreover, human visual perception and processing are heavily influenced by top-down cognitive control and prior knowledge \cite{failing2018selection} that allocates human attention to task-relevant objects~\cite{deng2016does,lim2009modeling,guo2021machine}. 
Such selective focus of attention to task-relevant objects hence supports human decision-making \cite{spering2022eye,wispinski2020models}. 
For instance, experienced drivers can easily identify and attend to task-relevant information (e.g., cars  or pedestrians in the front) for driving performance and ignore salient but irrelevant information. 
The reciprocal link among eye movements, visual perception, attention, and decision-making in humans suggests that an AIV needs to be trained with a sophisticated reasoning mechanism similar to that of humans. 
Motivated by this, prior works \cite{xia2019predicting,fang2021dada,palazzi2018predicting} have aimed to model drivers' visual attention in driving contexts, suggesting where drivers need to attend. 
However, the integration of this human attention prediction system into autonomous driving systems to help AIV decision-making and reasoning receives scant attention.
Hence, it remains an open question whether using human attention as training guidance for machine
decision-making has an effect on reinforcement learning (RL) for AIVs.

Recent decision-making methods for AIVs have not incorporated human attention and have mainly adopted two end-to-end deep learning approaches: imitation learning (IL) \cite{pomerleau1988alvinn} and deep reinforcement learning (DRL) \cite{RN40}. 
IL aims to learn driving strategies from an expert, such as a human driver, by mimicking their control actions in similar situations \cite{pomerleau1988alvinn, chib2023recent,codevilla2018end,wen2021keframe,inproceedingsgaze,shao2023reasonnet}. 
However, IL faces a notable limitation: its vulnerability to the distribution shift problem \cite{chang2021mitigating}. 
During training, an IL model learns from a specific distribution of states and actions from the expert. 
This means that IL models usually do not explore sufficiently in scenarios where unforeseen failures occur, hindering their ability to respond correctly under adverse conditions~\cite{toromanoff2020end, filos2020can, zheng2022imitation, yu2023offline}.
Unlike IL, DRL mitigates the distribution shift problem because it enables agents to learn through trial and error by rewarding chosen actions and allowing them to adapt to new environments \cite{mnih2013playing}. 
However, DRL has two limitations. One major limitation of DRL is the substantial data and time requirements for convergence. 
This is due to the sparse reward signals and the RL agent needing extensive exploration to learn effective policies \cite{kumar2019stabilizing}. 
Another limitation of DRL, which it shares with IL, is the lack of explainability inherent in the deep neural networks used by both approaches. 
These networks map perception to actions in an opaque manner, making the decision-making process of machines nontransparent~\cite{zablocki2022explainability, ou2023fuzzy}.


Current AIV models have been reported to be involved in several accidents~\cite{adadi2018peeking, stanton2019models, atakishiyev2024explainable}. 
The lack of interpretability in these models has raised public concerns about the need for explainable decision-making systems~\cite{ali2023explainable} and even calls for legislation~\cite{AIAct2024}. 
This lack of transparency makes it difficult to understand the decisions that lead to these worrying accidents. 
Unlike earlier approaches focusing mainly on performance, our method emphasizes human-centric computing with a focus on interpretability.
 Interpretability is the ability to explain or provide meaning in terms understandable to humans~\cite{arrieta2020explainable}. 
 The interpretability of our framework has two main aspects. 
 First, aligning machine attention with human attention makes the model's behavior more understandable to humans. 
 Human attention reflects how people focus on important information, and with proper guidance, visual search can be influenced by top-down cognitive control~\cite{wolfe2010visual}. 
 Second, our network uses a self-attention mechanism to represent machine attention. 
 This enhances interpretability by helping the model learn to focus on the most important features of the input~\cite{mascharka2018transparency}.
Additionally, we conducted a survey of 80 respondents (see Fig.~\ref{fig:q_human}), which shows that our model’s attention and decision-making process are easier to understand than those of existing approaches.

This paper introduces Human Attention-based Explainable Guidance for Intelligent Vehicle Systems (AEGIS) as a solution to the interpretability issue and a response to the open question of the effectiveness of human attention guidance in RL for AIVs.
In contrast to previous human-guided RL approaches \cite{wu2023toward,wu2022prioritized,wu2023human} that have primarily provided action guidance, AEGIS leverages human attention to guide the RL agent on the latent code of action. 
The proposed attention-based guidance enables the RL agent to learn task-relevant objects, thereby improving its generalizability (see Tab. \ref{tab:supp_headings} and Tab. \ref{tab:supp_drop-left-turn}). 
To acquire human drivers' attention, we collected large-scale eye-tracking data using a realistic VR driving simulator (see Sec. \ref{sec:dataset}). 
We recorded the active engagement of drivers and propose a unique framework that incorporated these human attention data into the training of DRL for autonomous driving tasks. 
Our dataset includes 20 participants engaged in two challenging scenarios (see Fig. \ref{fig:supp_fig1} and Fig. \ref{fig:supp_fig2}) and four diverse occlusion scenarios inspired by~\cite{shao2023reasonnet} (see Fig. \ref{fig:supp_occ}), yielding a total of six scenarios, 1.2 million frames and 600 minutes of driving data.
To the best of our knowledge, this is the largest eye-tracking dataset collected using an immersive method with a VR headset and physical simulator (see Fig.~\ref{fig:supp_fig1} and Tab.~\ref{table:supp_dataset}).
Leveraging this dataset, we craft an explainable driving model that utilizes human attention predicted from a model pre-trained on the eye-tracking dataset to guide the model's self-attention layer.
The pre-trained human attention model, while simplistic, eliminates the need for eye-tracking data during inference. 
Compared with the traditional RL without human attention integration, our method enhances focus on crucial objects and increases the DRL training speed (see Fig.~\ref{fig:teaser}). 
Moreover, AEGIS ensures the similarity of machine attention with human attention, thereby increasing agent robustness in new scenes. 
Although AEGIS prioritizes explainability, which is an important topic for AI safety in both research and industry, performance analysis and attention visualization confirm that integrating human attention significantly benefits agents.
The contributions of this work are four-fold:
\begin{itemize}
\item A novel and largest in-lab eye-tracking dataset collected using a realistic VR driving simulator, capturing drivers' active engagement across six diverse scenarios.
\item The incorporation of human attention guidance aligns machine attention more closely with learned human attention, improving the RL agent's explainability.
\item The proposed AEGIS framework involves a human-attention guidance mechanism to enable the RL agent to learn task-relevant objects.
\item Comprehensive analysis shows that AEGIS significantly improves training efficiency, robustness in unseen scenes, and overall performance. 
\end{itemize}
\section{Related Work}
\label{sec:rel_work}

\subsection{Human action-guided RL}
RL has shown great success in complex tasks such as playing games that can surpass human players in Atari~\cite{mnih2013playing} and Go \cite{silver2017mastering}. 
However, the training efficiency of RL is hindered by its requirement for extensive interactive sessions and a propensity to converge on suboptimal solutions due to insufficient prior knowledge~\cite{wu2023human}. 
Prior works \cite{luo2023human,wu2023toward,wu2022prioritized,wu2023human} have attempted to increase the sample efficiency and performance of RL via human guidance, which provides necessary prior knowledge with human demonstrations to the RL during training. 
Suboptimal action replacement \cite{luo2023human,wu2023toward,wu2022prioritized,wu2023human}, reward shaping \cite{wu2022prioritized, li2022efficient}, and replay buffer prioritization \cite{wu2023toward,wu2022prioritized,wu2023human} have been proposed to accelerate RL training. 
In human-in-the-loop RL frameworks with suboptimal action replacement \cite{luo2023human,wu2023toward,wu2022prioritized,wu2023human}, humans can intervene and replace RL actions with their actions. 
This is based on the assumption that humans can correct suboptimal RL behaviors when necessary. 
By doing so, training speed and performance could be improved. 
Reward shaping is another common method in human-in-the-loop RL, where a negative reward penalizes RL when human actions have deviated from RL actions \cite{wu2022prioritized, li2022efficient}. 
Wu et al. \cite{wu2023toward,wu2022prioritized,wu2023human} used prioritized experience replay mechanisms \cite{schaul2015prioritized} to prioritize human demonstrations based on Q value difference between human and RL actions. 
However, these studies relied on human actions to replace suboptimal RL agent actions and required substantial human demonstrations, and depended that humans remain available throughout the RL training process.
Moreover, these works still leave the decision-making process a black box. 
Unlike these works, we provide human attention knowledge to the latent space of the RL agent via the attention mechanism and thus enhance the interpretability of our RL agent.

\subsection{Machine attention of RL}

Extensive research has been conducted to explain the black box behavior of the neural network of the RL agent \cite{zhou2016cvpr, wang2020score, DBLP:journals/corr/abs-2010-03023,greydanus2018visualizing, zambaldi2018deep, mott2019towards, tang2020neuroevolution, itaya2021visual, nikulin2019free}. 
Joo et al.\cite{joo2019visualization} employed the Gradient-weighted Class Activation (Grad-CAM) \cite{selvaraju2017grad} method to explain the area of an image that is related to the decision. 
Similarly, Greydanus et al.\cite{greydanus2018visualizing} showed that the Atari agent could be explained via an occlusion-map method. 
Nevertheless, these post-hoc explanation methods can be computationally expensive. 
For example, the occlusion-map approach outlined in \cite{greydanus2018visualizing}, requires 256 model inferences to explain a single $80 \times 80$ Atari game image, rendering it impractical for real-time applications. 
Additionally, as these explanations are generated after model training, they are often approximations and may overlook critical aspects of the model's decision-making process \cite{mhasawade2024understanding}. 

An alternative approach to achieve interpretability in RL involves the development of inherently transparent models. 
For instance, Zambaldi et al. \cite{zambaldi2018deep} developed a relationship module that leveraged the self-attention mechanism \cite{vaswani2017attention} to explain the focus area in the Starcraft II environment. 
Similarly, the self-attention mechanism was applied in \cite{mott2019towards, tang2020neuroevolution, itaya2021visual, nikulin2019free} to augment agent representation and interoperability. 
Unlike these works, the self-attention layer of RL policy network in our approach is guided by human attention and explored in AIVs.

\subsection{Human attention for RL}

Unlike machine attention, human attention can integrate diverse sensory inputs into a coherent understanding and involve sophisticated cognitive processes, such as attention allocation, that contribute to decision-making \cite{lindsay2020attention, lai2020understanding}.
While some imitation learning (IL) studies~\cite{zhang2018agil,saran2020efficiently,zhang2020atari, chen2019gaze, zhang2018learning, zhang2017attention} have incorporated human attention, to the best of our knowledge, no RL research has investigated the integration of human attention guidance within an RL framework for visuomotor control tasks, such as playing Atari games. 
Notably, one RL work~{\cite{guo2021machine}} compared human and machine attention in RL agents for Atari games but did not integrate human attention in the RL framework.

In driving studies, various attempts have been made to predict the drivers' attention in in-car \cite{palazzi2018predicting, alletto2016dr} and in-lab datasets \cite{xia2019predicting,fang2021dada, gopinath2021maad, deng2023driving, Baee_2021_ICCV}. 
However, the in-car dataset cannot collect repeated scenarios for different drivers, and the in-lab dataset often requires the driver to review a video without active engagement of the task. 
To solve these issues, we employ a realistic VR simulator for active engagement and realistic data collection. 
With these datasets, several convolutional neural networks (CNNs) \cite{xia2019predicting,fang2021dada,palazzi2018predicting} have been proposed to predict drivers' attention.
Despite these attempts, a notable gap remains in applying human visual attention as the prior knowledge for training the RL agent in driving tasks. 
Unlike these works which focused on accurately predicting gaze position, our work focuses on exploring the effect of learned human attention on the RL framework for the control process of AIVs.



\section{Dataset}
\label{sec:datasets}

\subsection{Eye-tracking data collection}
\label{sec:dataset}
\textbf{Author Statement:}
As the authors of this dataset, we take full responsibility for its integrity and any issues related to data rights or ethical standards. 
We confirm that the collection and use of data comply with relevant regulations, and all participants were compensated at an hourly rate higher than the country's minimum wage.
The dataset is shared under an MIT license, allowing use, redistribution, and citation that align with the license term.\\
\begin{figure}[htb!]
\centering
\subfloat[Data collection system]{%
  \includegraphics[width=0.8\linewidth]{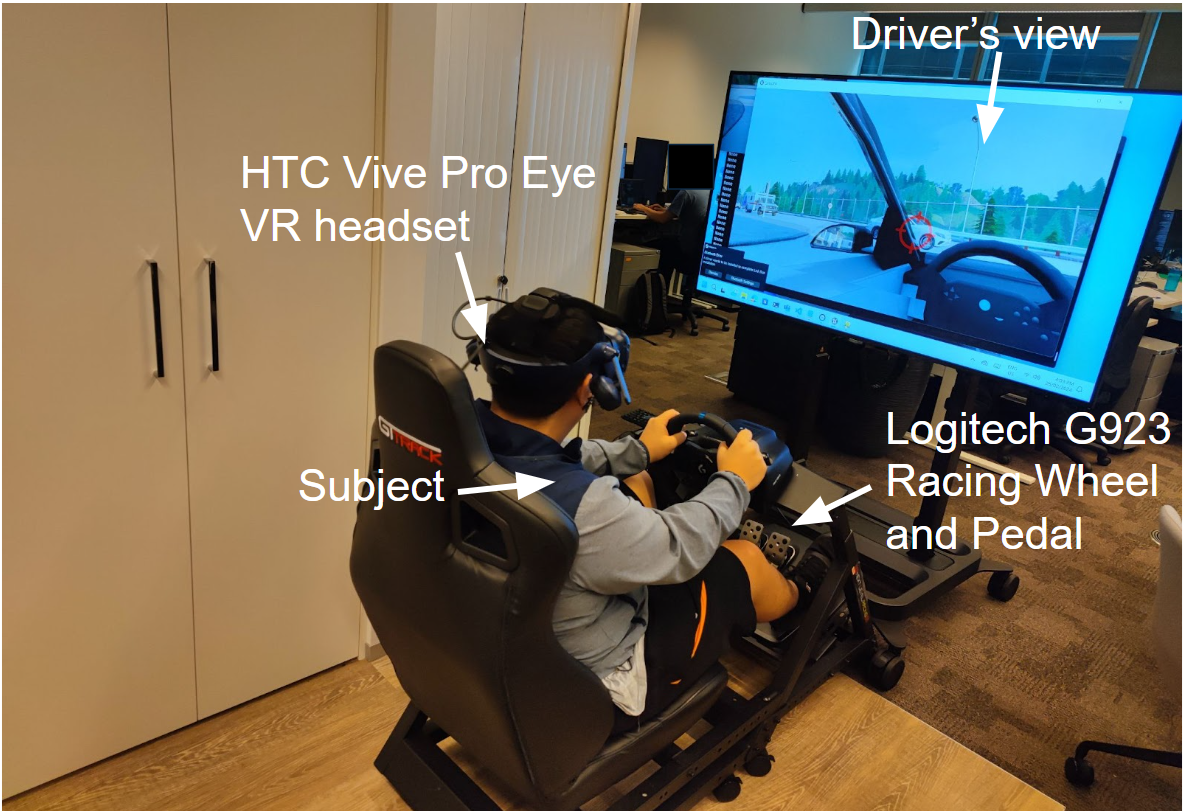}%
  \label{fig:sub-first}
}
\hfill
\subfloat[Views]{%
  \includegraphics[width=0.8\linewidth]{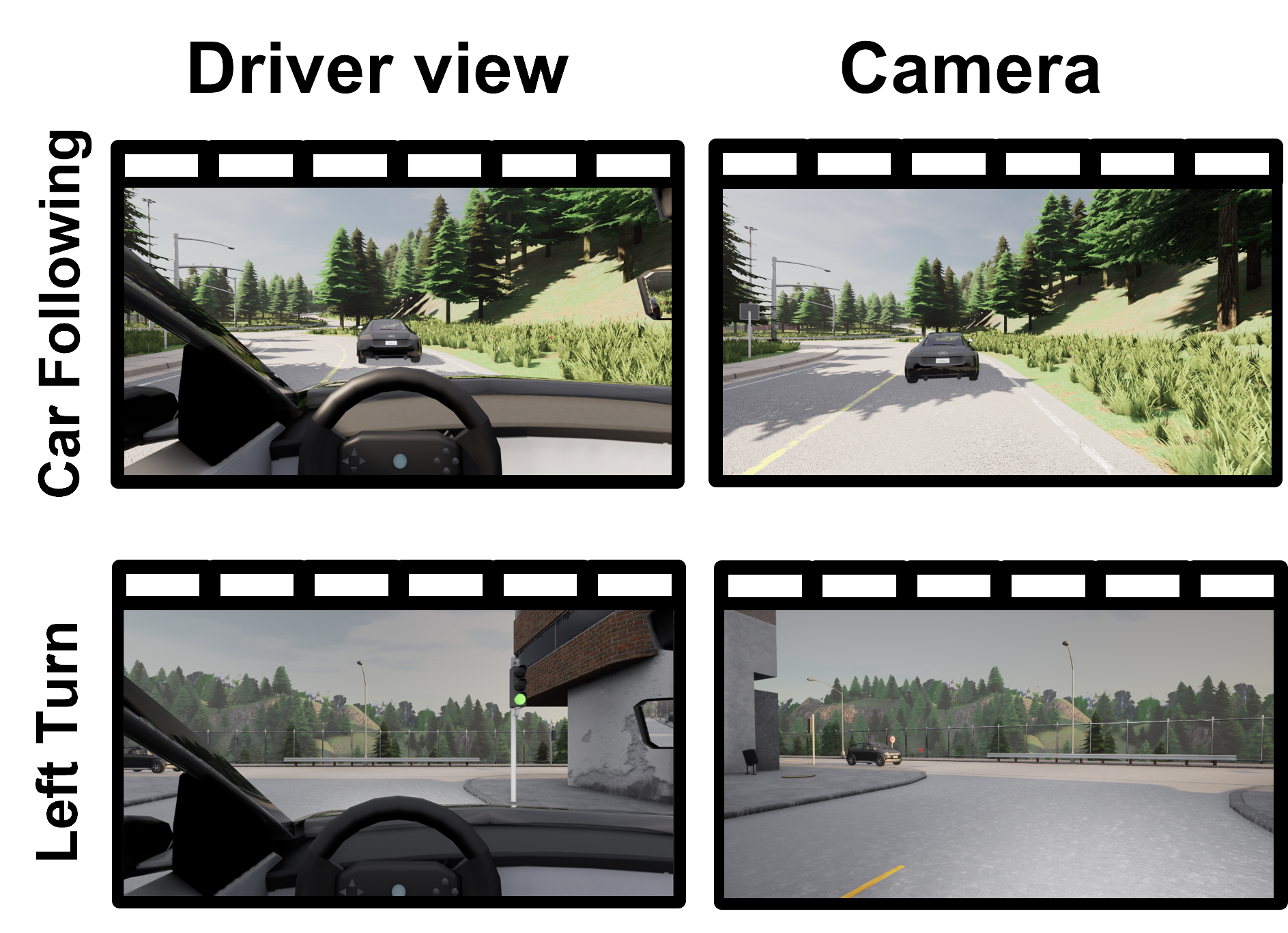}%
  \label{fig:sub-second}
}
\caption{The dataset collection environment. The HTC VIVE Pro Eye VR headset and Logitech G923 Racing Wheel and Pedal give the subject a more realistic driving experience.}
\label{fig:supp_fig1}
\end{figure}
\textbf{Dataset Description:}
To simulate a realistic driving experience, we integrate VR technology using an HTC VIVE Pro Eye VR headset (resolution: $2\times1440\times1600$, refreshing rate: 90 Hz, eye-tracking accuracy: 0.5$^{\circ}$-1.1$^{\circ}$) and an open-source CARLA simulator \cite{dosovitskiy2017carla, silvera2022dreyevr}. 
The study includes 20 participants with normal or correct-to-normal vision (age: $24.1 \pm 4.8$ years) who were provided informed written consent before participation. 
The participants have an average driver's license possession of $4.9 \pm 3.9$ years. 
The participants filled out the driver skill inventory (DSI) form \cite{lajunen1995driving} prior to the experiment, with a perceptual-motor skills score of $4.1 \pm 0.3$ and a safety skills score of $3.7 \pm 0.3$ (scale: 1-5).
The university's human research ethics committee has approved the protocol for involving human participants. 
The participants signed the consent form prior to the experiment, and were fully aware of the purpose of the data collection.
The participants were instructed to complete the driving tasks to the best of their ability, with their eye movements being recorded via the VR headset. 
Each driving scenario lasts about five minutes, with six diverse scenarios in total, and the total dataset collection time is 600 minutes. The dataset allows training of a human attention model, and no further data collection is required in RL training.

\begin{table*}[h!]
\caption{Comparison between AEGIS dataset and other eye-tracking driving datasets, updated according to \cite{kotseruba2022attention}. Our eye-tracking dataset is the largest dataset, adopting a realistic VR driving simulator. The following abbreviations are used in the table. Camera: S - scene facing camera, RGB - 3-channel image, sem - semantic segmentation mask. Frame counts marked with * are estimated based on the lengths of the videos and camera frame rate.}
\centering
\begin{tabular}
{@{}l@{}c@{\hspace{5pt}}c@{\hspace{8pt}}c@{\hspace{8pt}}c@{\hspace{8pt}}c@{\hspace{8pt}}c@{\hspace{8pt}}c@{\hspace{5pt}}c@{\hspace{5pt}}c}
\toprule
Dataset  & Active control & Vehicle data & Camera   & Hazards & View & \#subjects & \#frames  \\ 
\midrule
\rowcolor{yellow}AEGIS (Ours) & + & + & $S^{rgb,depth,sem}$  & + & VR & 20 & \textbf{1.2 M} \\ 
DrFixD(night) \cite{deng2023driving} & - & - & $S^{rgb}$ & - & screen & 31 & 67K* \\ 
LBW \cite{kasahara2022look} & - & - & $S^{rgb,depth}$ & - & screen & 28 & 123K* \\ 
CoCAtt \cite{shen2022cocatt}  & + & + & $S^{rgb}$ & - & screen & 11 & 17K* \\ 
MAAD \cite{gopinath2021maad} & - & - & $S^{rgb}$  & - & screen & 23 & 60K \\
TrafficGaze \cite{deng2019drivers} & - & - & $S^{rgb}$ & - & screen & 28 & 77K* \\ 
DADA-2000 \cite{fang2019dada} & - & - & $S^{rgb}$ & + & screen & 20 & 658K  \\ 
DR(eye)VE \cite{alletto2016dr} & + & - & $S^{rgb}$ & - & on-road & 8 & 555K \\ 
BDD-A \cite{xia2019predicting} & - & + & $S^{rgb}$ & + & screen & 45 & 378K* \\ 
C42CN \cite{taamneh2017multimodal} & + & - & $S^{rgb}$ & + & screen & 68 & - \\ 
TETD \cite{deng2016does}  & - & + & $S^{rgb}$ & - & screen & 20 & 100\\
3DDS \cite{borji2011computational} & + & - & $S^{rgb}$ & - & screen & 10 & 192K \\ \bottomrule
\end{tabular}

\label{table:supp_dataset}
\end{table*}

In the experiment, participants could control the vehicle through a racing wheel and pedals, closely mimicking real-life driving (see Fig.~\ref{fig:supp_fig1}). 
This arrangement enables us to collect accurate eye-tracking information while ensuring participants are actively involved. 
Moreover, it permits the replication of the same scenarios with the same initial settings and exo agent configurations for different individuals to develop a model that can be applied across individuals.
Our data collection pipeline represents a significant advantage by combining scenario-level replication, immersive VR over 2D screens, and active participant control for more realistic and consistent data collection.
These represent a distinct advantage over previous eye-tracking driving datasets (see Tab.~{\ref{table:supp_dataset}}). 

To date, the only large on-road dataset is the DR(eye)VE \cite{alletto2016dr} dataset. 
However, the major limitation of the on-road dataset is that the traffic conditions and exo agents' behaviors are not replicable for different drivers \cite{kotseruba2022attention}. 
Several studies \cite{gopinath2021maad, xia2019predicting} have leveraged this dataset and collected eye-tracking data from experienced drivers by asking them to watch the driving videos and imagine as if they are driving. 
However, the previous approach results in passive observation, which alters gaze distribution compared with actual driving due to the lack of vehicle control and task mindset by participants \cite{xia2019predicting}. 
Moreover, screen-based displays offer a limited field-of-view (FOV) and a less realistic driving experience.
Similarly, 3DDS~\cite{borji2011computational} , C42CN~\cite{taamneh2017multimodal}, DADA-2000~\cite{fang2019dada}, LBW~\cite{kasahara2022look}, DrFixD(night)~\cite{deng2023driving}, and CoCAtt~\cite{shen2022cocatt} record gaze data in a low-fidelity screen-based driving simulator. 
To solve the issue of screen-based simulators, we collect our AEGIS dataset, the first large eye-tracking driving dataset recorded from a VR driving simulator. 
This setting can provide a $360^\circ$ FOV with much higher fidelity to real driving \cite{kotseruba2021behavioral}. 
Additionally, our dataset includes 1.2M frames collected from 20 participants, with the location of their eye gaze recorded on each frame. 
These frames are recorded from three cameras, including RGB, semantic, and depth cameras. 
In addition to visual data, our dataset includes comprehensive vehicle control information, such as throttle and brake inputs, as well as vehicle dynamics data (e.g., speed and acceleration metrics). 
Similar to DADA-2000 and BDD-A \cite{fang2019dada,xia2019predicting}, we focus on critical scenarios that require immediate decision-making. 
We then train a human attention network with the dataset (see Sec.~\ref{human_att_net}).


\subsection{Scenario design}

\begin{figure}[tbh!]
  \centering
  \includegraphics[width=\linewidth]{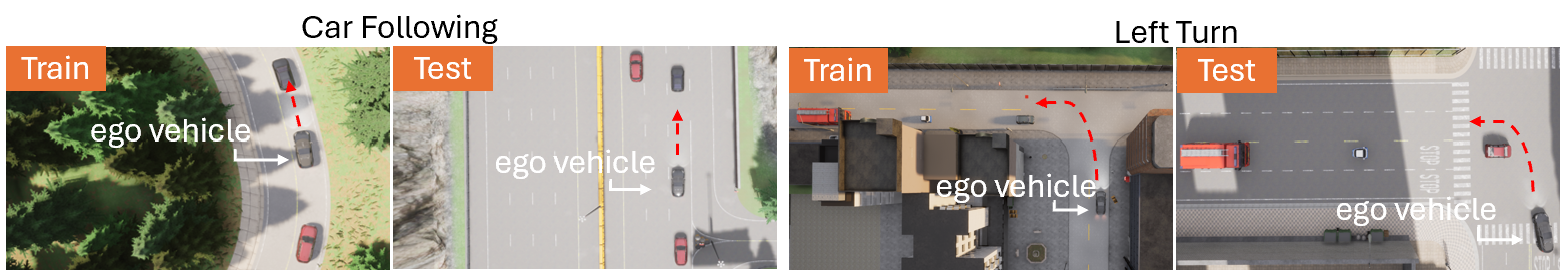}
  \caption{Car Following: The ego vehicle must avoid collisions with the car ahead by controlling the throttle and brake, ensuring it continues to follow the lead car. Left Turn: The ego vehicle must accurately time its left turn to avoid collisions with vehicles proceeding straight by controlling the throttle and brake.}
  \label{fig:supp_fig2}
\end{figure}

\begin{figure}[tbh!]
  \centering
  \includegraphics[width=\linewidth]{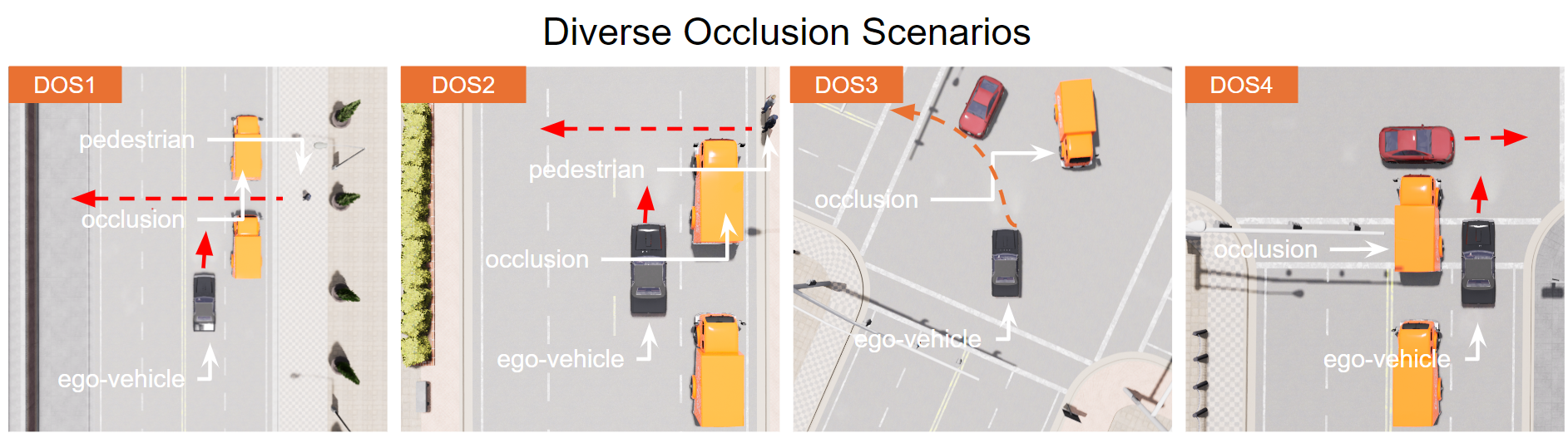}
  \caption{Diverse occlusion scenes. The ego vehicle must control the throttle and brake to prevent collisions with occluded objects, such as pedestrians and cars.}
  \label{fig:supp_occ}
\end{figure}

\begin{figure}[htb!]
  \centering
\includegraphics[width=\linewidth]{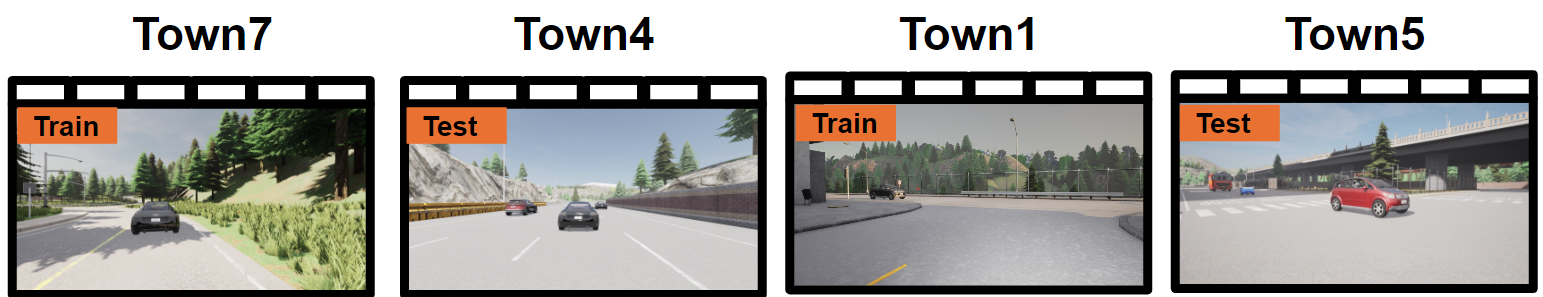}
  \caption{Training and Testing scene. The car-following model is trained in Town 7, characterized by its rural setting and narrow roads, and then tested in Town 4, a mountainous area featuring highways. The left-turn model is trained in Town 1, a small town, and then tested in Town 5, a town with bridge and cross junctions.}
  \label{fig:supp_towns}
\end{figure}

Our method is validated in six scenarios: car following, left turn (see Fig.~\ref{fig:supp_fig2}), and four diverse occlusion scenarios \cite{shao2023reasonnet} (see Fig. \ref{fig:supp_occ}). The RL agent is trained and evaluated in different towns (see Fig. \ref{fig:supp_towns}) to enhance the challenge and diversity, so the agent must develop a robust representation to avoid collisions in unseen scenes. A PID controller controls the lateral movement following \cite{wu2022prioritized}, while the RL agent concentrates on throttle and brake control, with its action $a_t \in [-1, 1]$.\\
\textbf{Car-following scenario: }
Inspired by \cite{wang2022velocity}, our car-following task resembles the adaptive cruise
control (ACC) for the traffic congestion situation. 
The ego vehicle must follow a lead vehicle within the same lane, which moves at speeds up to 8 m/s and may brake abruptly. 
The ego vehicle should brake swiftly to avoid collisions while maintaining a close enough distance for effective following.
This scenario design is challenging, and many participants found it harder than the left-turn scenario and some participants stated that the car-following scenario demanded sustained attention over a longer duration to monitor the lead vehicle.\\
\textbf{Left-turn scenario:}
Adopting from \cite{wu2022prioritized}, our left-turn scenario requires the ego vehicle to perform a left turn at an intersection, ensuring no collisions with oncoming vehicles. These vehicles, moving at speeds between 3m/s and 5m/s, act aggressively and do not give way to the ego vehicle. The ego vehicle should blend into the traffic at the right moment, aiming to reach a goal point swiftly. 
\\
\textbf{Occlusion scenarios:}
The four occlusion scenarios follow the public Drive
in Occlusion Simulation (DOS) benchmark \cite{shao2023reasonnet}. The benchmark consists of 100 diverse cases with oncoming vehicles or pedestrians occluded by other vehicles. DOS1 and DOS2 involve scenarios where a pedestrian, occluded by cars, is walking across the road. In DOS1, the ego vehicle can avoid a collision by detecting the pedestrian early, before the occlusion occurs. In DOS2, the ego vehicle must adapt by reducing its speed when approaching the intersection, as the cars completely obscure the pedestrian from view beforehand. In DOS3, the ego vehicle should slow down when passing the intersection to ensure safety. In DOS4, the ego vehicle can identify the oncoming traffic through the gaps between the obstructing vehicles.

\section{Methods}
\label{sec:methods}
\subsection{RL problem definition}
\label{RL_Def}

Our task is a goal-oriented collision-avoidance task that involves controlling the ego vehicle's brake and throttle strength. 
The Markov Decision Process (MDP) for this task can be represented as $\{S, A, P, R\}$. 
At a time step $t$, the agent (e.g., ego vehicle) observes the state $s_t$ from all possible states $S$ and outputs an action $a_t$ from the action space $A$.
Given this action, the environment transitions to a new state $s_{t+1} \in S$ according to the transition probability matrix $P$ and provides a reward $r_t$. 
The reward is determined by the reward function $R(\cdot | s, a): S \times A \to r$.
The goal of reinforcement learning (RL) is to find an optimal policy $\pi$ that maximizes the expected return $\sum_{k=0}^{\infty} \gamma^k r_{t+k}$, where $\gamma \in (0,1)$ is the discount factor.
\\
\textbf{State space: }
The state space $ S = \big\{ s_t \mid s_t = \{I_{t-2}, I_{t-1}, I_t\}, \, t \in \mathbb{N} \big\} $ 
is defined as the set of all sequences of three consecutive segmentation images, where 
$ I \in \mathbb{R}^{h \times w \times 1} $ represents a single segmentation image captured 
by the camera sensor in the CARLA simulator~{\cite{dosovitskiy2017carla}}.
\\
In the car-following scenario, the camera faces forward. 
For the left-turn scenario, the looking vector is adjusted 30$^{\circ}$  to the left to mimic a driver's perspective. 
In the occlusion scenarios, we employ three cameras with a front-facing camera and two side cameras facing 60$^{\circ}$  to the left and right due to pedestrians coming from the sidewalk.  \\
\textbf{Action space: }
The action space $A = \{ a_t \mid a_t \in [-1, 1], \, t \in \mathbb{N} \}$ represents the set of possible longitudinal control commands for the ego vehicle.
 A value of $-1$ indicates the maximum braking force, whereas $1$ denotes the maximum throttle force. \\
\textbf{Reward: }
The primary objective of the RL agent is to avoid collision while completing tasks efficiently. To encourage such behavior, we design a reward function as follows:
\begin{equation}
\begin{split}
  r_t &= R(\cdot|s_t,a_t) \\
  &= r_{goal}(s_t \in C_{goal}) + r_{collide}(s_t \in C_{collide})\\
  &\quad + \omega*r_{idle}(s_t \in C_{idle}) + \delta*r_{gap}(s_t)
  \label{eq:rlt}
\end{split}
\end{equation}
where $C_{goal}$, $C_{collide}$, $C_{idle}$ and $r_{gap}$ represent completion, collision, idle status, and time gap, respectively.
The agent is awarded a large positive reward $r_{goal}=100$ when it reaches the goal and receives a large negative reward $r_{collide}=-100$ when a collision happens, a negative reward $r_{idle}=-1$ when the ego vehicle stops moving. The time gap is the time it takes for the ego vehicle to reach the current position of the lead vehicle.
Similar to~{\cite{dagdanov2023self}}, we define a reward term $r_{gap}$ for the car-following scenario is as follows:
\begin{equation}
  r_{gap}=\begin{cases} 
   T_{gap} & \text{if } T_{gap} \in [1,2] \\
   \max(-1/T_{gap},-10) & \text{if } T_{gap}<1 \\
  \max(-T_{gap}, -10) & \text{if } T_{gap}>2
\end{cases}
\end{equation}
where $T_{gap} = Dis/V_{ego}$, where 
$Dis$ is the distance between the ego vehicle and the lead vehicle, and $V_{ego}$ is the speed of the ego vehicle. This reward design encourages the ego vehicle to maintain a safe and optimal distance from the lead vehicle.

\subsection{Policy network}
\begin{figure*}[htb!]
  \centering
  \includegraphics[width=\linewidth]{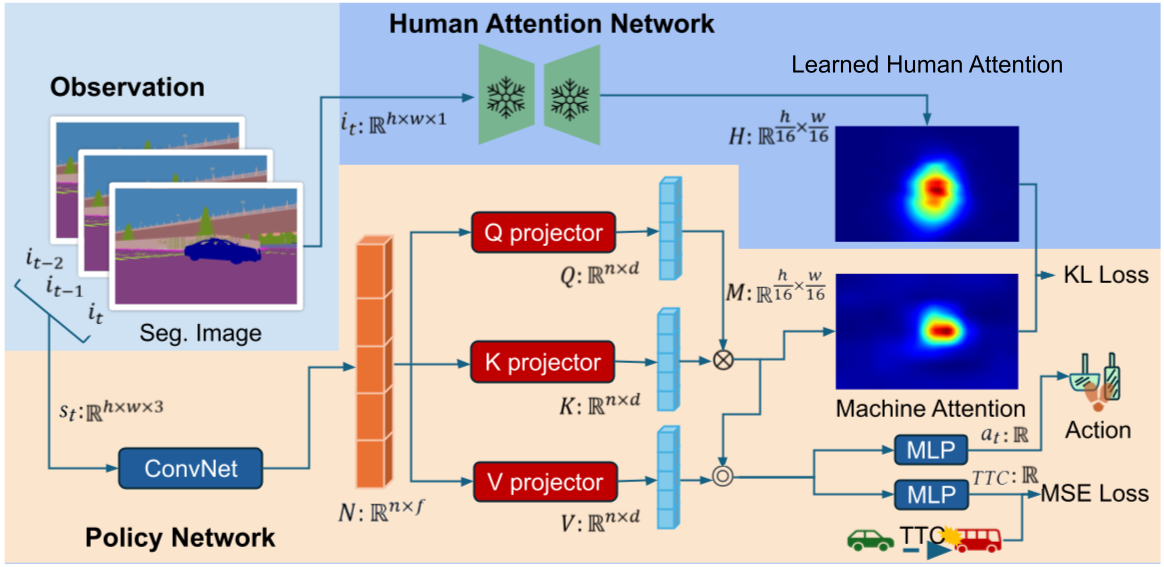}
  \caption{ Structure of AEGIS. \textbf{Human Attention Network:} This pre-trained network predicts human attention from a segmentation image. \textbf{Policy Network:} This network determines the vehicle's policy from a sequence of three segmentation images, starting with a CNN to extract features. These features are then flattened and processed through a self-attention layer, producing machine attention. This machine attention regulates RL training using the KL divergence loss relative to human attention. The policy network includes two MLP prediction heads: one for estimating the throttle and brake strength and another for predicting TTC, which aids in training regularization through the MSE loss. $\otimes$ represents dot product, and $\circledcirc$ represents scaled dot product (after normalization and Softmax). 
}
  \label{fig:fig3}
\end{figure*}
We design an interpretable policy network leveraging the self-attention mechanism (see Fig.~\ref{fig:fig3}). 
The model processes three consecutive semantic segmentation images as the input state
$s_t \in \mathbb{R}^{h \times w \times 3}$, where $h$ and $w$ denote the height and width of the input images, respectively. 
The policy network outputs machine attention $M\in $ $\mathbb{R}^{h/16 \times w/16}$, an action $a_t \in [-1, 1]$, which represents throttle and brake control, and time-to-collision (TTC) within the range $[0, 5]$, in a unified framework. 
A shallow CNN encodes the semantic segmentation images into a feature map $F \in \mathbb{R}^{h/16 \times w/16 \times f}$.
This feature map is subsequently flattened into $N \in \mathbb{R}^{n \times f}$, where $n = \frac{h}{16} \times \frac{w}{16}$.
The flattened representation is then projected into the query $Q$, key $K$, and value $V$ matrices via fully-connected layers $f_Q$, $f_K$, and $f_V$, respectively:
\begin{equation}
Q = f_Q(N), \quad K = f_K(N), \quad V = f_V(N)
\end{equation}
Here, $Q, K, V \in \mathbb{R}^{n \times d}$, where $d$ represents the dimensionality of the latent space for each token.
These matrices are utilized to compute self-attention as follows:
\begin{equation}
\begin{split}
\text{SelfAttention} &= \text{softmax}\left(\frac{QK^\top}{\sqrt{d}}\right)V \\
&= \text{MachineAttention}(Q, K)V \\
&= MV
\label{eq:attention}
\end{split}
\end{equation}
In this formulation, $QK^\top \in \mathbb{R}^{n \times n}$ encodes the pairwise attention scores between all the elements, which are normalized by $\sqrt{d}$ to improve numerical stability.
The attention scores are normalized using the softmax function, converting them into a probability distribution where the weights sum to 1 for each element. 
The resulting matrix $M \in \mathbb{R}^{n \times n}$, referred to as machine attention, determines the relative importance of each element. 
These attention weights are applied to $V$, producing the final output of the self-attention mechanism.

The output from the self-attention layer is flattened and inputs into multilayer perceptrons (MLPs) to predict action $a_t \in [-1, 1]$ where $-1$ represents maximum braking, and $1$ indicates maximum throttle, and TTC:
\begin{equation}
  TTC = clip(Dis / (V_{ego} - V_{front}), 0, 5) 
  \label{eq:ttc}
\end{equation}
where $Dis$ is the distance to the closest vehicle, $V_{ego}$ is the speed of the ego vehicle, and $V_{front}$ is the speed of the closest vehicle. We clip $TTC$ to the range $[0,5]s$ to encourage the agent to concentrate on critical situations. 

 To regularize learning, we employ Kullback-Leibler Divergence (KL) $\mathcal{L}_{kl}$ to align machine attention with learned human attention from the pre-trained model in Sec.~\ref{human_att_net}, and the mean square error $\mathcal{L}_{mse}$ to align the predicted TTC with the ground truth. 
Overall, the new loss is:
\begin{equation}
  \mathcal{L}_{total} = \mathcal{L}_{\pi} + \alpha * \mathcal{L}_{kl} + \beta *\mathcal{L}_{mse}
  \label{eq:loss}
\end{equation}
where $ \mathcal{L}_{\pi} $ is the loss of the original RL policy network, which depends on the RL method used, and $ \mathcal{L}_{kl} $ and $ \mathcal{L}_{mse} $ are auxiliary losses used for regularization.
Notably, we only supervise the machine attention with learned human attention in the first 500 steps of RL training, allowing the agent to further refine its attention after that.


\subsection{Human attention network}
\label{human_att_net}

In this study, we utilize a CNN model developed in \cite{deng_CDNN}, which is based on a lightweight U-Net \cite{ronneberger2015u} structure.
To mitigate the variability and potential distraction arising from individual differences and ensure consistency in our results (see Fig.~\ref{fig:supp_eye_tracking}), we employ a human attention network that can predict the general pattern of humans' focus of attention. This choice is motivated by the model's balance of computational efficiency and speed, making it suitable for real-time applications. The inference time of one image is 0.005s. The ground truth of the model is the human attention obtained from the gaze position of the previous ten consecutive frames, similar to the method in~{\cite{palazzi2018predicting}}.
The discrete gaze positions are converted into continuous distribution via a 2D Gaussian filter with \(\sigma\) that is equivalent to one visual degree \cite{le2013methods}, which is the visual field of the foveola, the high-acuity region of the retina at the center of gaze \cite{poletti2017selective}.
The human attention model is trained using binary cross entropy (BCE) loss until converged.


\section{Results}
\label{sec:results}


\subsection{Experimental Settings}
In this section, we discuss the training and testing scene settings and their differences for all scenarios in \textit{Scene Settings}, present the metrics for benchmarks and attention maps in \textit{Evaluation Metrics}, and provide details of the baselines and the hyperparameters of AEGIS in \textit{Baseline Methods}.
\\
\textbf{Scene settings.}
We use CARLA \cite{dosovitskiy2017carla} to construct the learning and testing environment for AEGIS. 
CARLA is an open-source driving simulator that offers maps of 14 towns for urban driving simulation. 
In our setup, the car-following scenario uses Town 7 as the training scene and Town 4 as the testing scene (see Fig.~\ref{fig:supp_towns}). 
The left-turn scenario uses Town 1 as the training scene and Town 5 as the testing scene.
Although we use semantic segmentation masks as inputs, significant differences still exist between different towns. For example, bridges and cross junctions in Town 5 represent new categories that are not present in the training scenes from Town 1. 
Additionally, variations in traffic conditions and road structure, such as narrower or broader roads, also influence performance and contribute to the differences between towns.
For four diverse occlusion scenarios, we follow the public Drive in Occlusion Sim (DOS) benchmark proposed in ReasonNet ~\cite{shao2023reasonnet}. We follow the training/evaluation setting in ReasonNet and use 5 cases for training and other 20 cases for evaluation for each scenario.
By testing in the unseen scenes, we can evaluate the models' generalizability in new environments. 
We train five models per method with different random seeds and report the average performance for a fair comparison. 
In the \textbf{free action setting}, each method can execute its own actions, which allows us to verify their performance. 
However, this setting leads to different observations from the environment within the same town. 
To compare visualization results across methods fairly, we also employ a \textbf{fixed action setting}, where the throttle value remains at 0.6 during inference.
\\
\textbf{Evaluation metrics.}
We evaluate the performance of the methods using the following metrics: a) Success rate, defined as the percentage of trials the agent reaches its destination without collision; b) Survival distance, which measures the distance traveled without any incidents; c) TTC as derived from Eq.~\ref{eq:ttc}; and d) Reward, calculated based on Eq.~\ref{eq:rlt}.
We evaluate the similarity between machine attention and human attention through the following distribution-based metrics: a) Pearson's Correlation Coefficient (CC); b) KL divergence; c) Similarity (SIM) \cite{swain1991color, bylinskii2018different}; and d) location-based metric Normalized Scanpath Similarity (NSS) \cite{peters2005components}. 
Our analysis goes beyond these metrics (see Sec. ~\ref{supp_focus_info_sec}).
\\
\textbf{Baseline methods.}
\textit{Benchmark for car-following and left-turn scenarios: }In these two scenarios, we adopt the TD3 algorithm \cite{fujimoto2018addressing} as the RL method for both AEGIS and all baseline methods. 
Specifically, we introduce a baseline referred to as Vanilla, which uses the same policy network architecture shown in Fig.~\ref{fig:fig3} but excludes the $\mathcal{L}_{\text{kl}}$ component. 
In addition to Vanilla, we include behavior cloning (BC)~\cite{pomerleau1988alvinn, hu2024thought}, a widely used approach in imitation learning (IL), as another baseline method. 
For the BC baseline, we construct a model to mimic human policy using the same network structure as the policy network in AEGIS. 
The model is trained with MSE loss on collected human demonstration data and then fine-tuned using RL. 
Unlike Vanilla, which starts training from scratch without prior knowledge, the BC baseline begins with a pretrained human policy and further refines it through RL.
This approach allows the BC model to incorporate human guidance directly in the action space.
By comparing AEGIS with both Vanilla and BC, we aim to emphasize the interpretability of human attention guidance.

As for training details of RL, the hyperparameters of TD3 include a replay buffer capacity of 38,400 and a minibatch size of 16, ensuring efficient data utilization during training. 
Learning rates for the actor and critic networks are $5 \times 10^{-4}$ and $2 \times 10^{-4}$, respectively, with a learning rate decay of 0.995 per episode to stabilize adjustments over time. 
Exploration rates decrease from $0.5$ to $0.05$ to balance exploration with exploitation, and the discount factor $\gamma$ is 0.95.
The hyperparameter settings for the loss in Eq. \ref{eq:loss} are $\alpha = 0.05$ and $\beta = 0.1$. $\alpha$ is an regularization term that allows the RL to maximize the reward while minimizing the KL divergence between human and machine attention.
The hyperparameter settings for the reward in Eq. \ref{eq:rlt} are $\omega = -1$ and $\delta = 1$.
All baseline methods use segmentation masks as input.

\textit{Benchmark for occlusion scenarios:} In these four scenarios~\cite{shao2023reasonnet}, we adopt the PPO algorithm~\cite{DBLP:journals/corr/SchulmanWDRK17} as the RL method for both AEGIS and all RL-based baseline methods to demonstrate the compatibility of the proposed framework with different RL methods.
We benchmark AEGIS with different attention guidance, including Class Activation Maps~\cite{zhou2016cvpr} (CAM-guided) and random masks (Random-guided) by replacing human attention.
By comparing AEGIS with Random-guided, we aim to demonstrate the advantages of focusing on task-relevant objects. 
Similarly, comparing AEGIS with CAM-guided allows us to assess the benefits of incorporating human guidance.
We also compare the result with ReasonNet~\cite{shao2023reasonnet}, a recent imitation learning method with cameras and LIDAR sensor fusion for a more comprehensive benchmark. 
We choose ReasonNet since it \textbf{ranks 1st} in the CARLA Leaderboard 1.0. 

For the training details of RL, the learning rate is $3 \times 10^{-4}$. 
The value function's importance is weighted by a coefficient of $0.5$, and a clipping range of $0.2$ helps maintain policy stability. 
The hyperparameter settings for the loss in Eq. \ref{eq:loss} are $\alpha = 0.05$ and $\beta = 0$ for occlusion scenarios.
The hyperparameter settings for the reward in Eq. \ref{eq:rlt} are $\omega = -0.2$ and $\delta = 0$.
Other hyperparameters of PPO are implemented using default settings in stable baselines 3~\cite{stable-baselines3}. All the baseline methods use segmentation masks as input, except for ReasonNet, which is a closed-source method that uses additional input such as LIDAR.

\begin{figure*}[tbh!]
  \centering
  \includegraphics[width=0.8\linewidth]{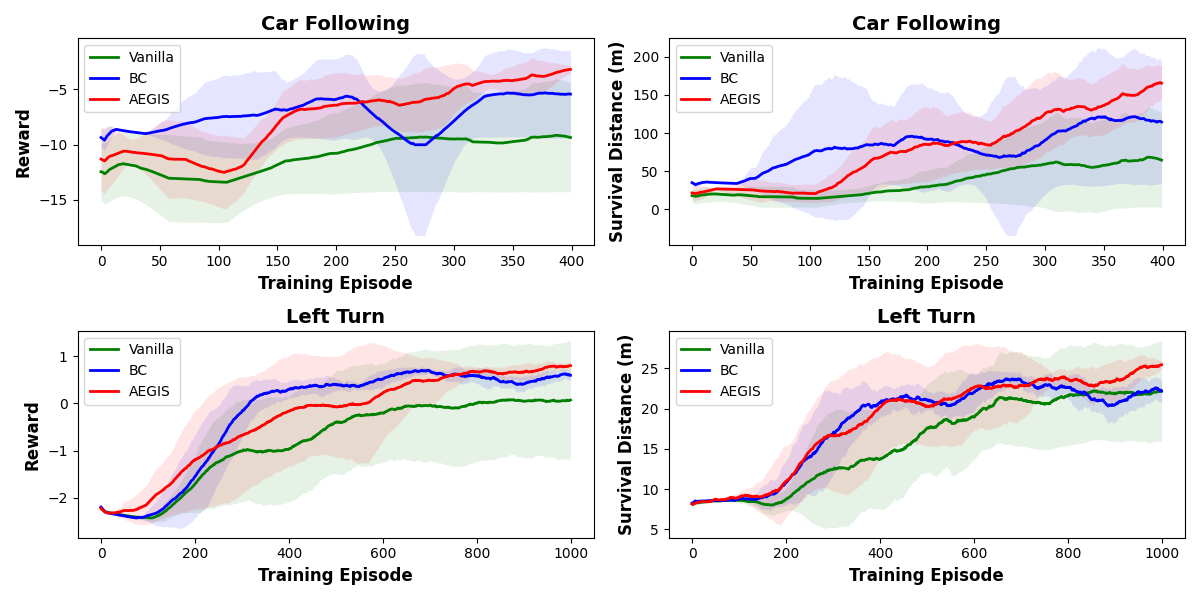}
  \caption{The training curves of AEGIS, Vanilla, and BC evaluated with reward and survival distances. With the help of learned human attention, AEGIS achieves the highest performance of Vanilla with fewer episodes.}
  \label{supp_fig:training_curve}
\end{figure*}




\begin{table*}[tb!]
  \caption{The evaluation results of car following / left turn in the unseen town with the free action setting. AEGIS outperforms Vanilla and BC.
  }
  \label{tab:tab1}
  \centering
\begin{tabular}{@{}l@{\hspace{10pt}}c@{\hspace{10pt}}c@{\hspace{10pt}}c@{\hspace{10pt}}c@{}}
    \toprule
    Model  & Success rate $\uparrow$ & Survival distance $\uparrow$  & TTC $\uparrow$ \\
    \midrule
    AEGIS (Ours) & $\mathbf{0.62 \pm 0.48}$ / $\mathbf{0.65 \pm 0.27}$ & $\mathbf{134\pm 72}$ / $\mathbf{18\pm 11}$ & $\mathbf{2.4 \pm 0.2}$ / $\mathbf{2.0 \pm 0.1}$\\
    BC & $0.46 \pm 0.40$ / $0.23 \pm 0.22$ & $97 \pm 76 $ / $13\pm 5$ & $2.3 \pm 0.1$ / $1.9 \pm 0.1$ \\
    Vanilla & $0.18 \pm 0.36$ / $0.33 \pm 0.36$ &  $35 \pm 41 $ / $15\pm 6$& $2.1 \pm 0.17$ / $1.8 \pm 0.1$\\ 

  \bottomrule
  \end{tabular}
\end{table*}
\begin{table}[b!]
  \caption{The success rate of AEGIS and baselines over four occlusion scenarios in the unseen town with a free action setting.}
  \label{tab:tab3}
  \centering
\begin{tabular}{@{}l@{\hspace{6pt}}c@{\hspace{6pt}}c@{\hspace{6pt}}c@{\hspace{6pt}}c@{}}
    \toprule
    Model  & DOS1    & DOS2     & DOS3   & DOS4 \\
    \midrule
    AEGIS (Ours) &  $\mathbf{0.66 \pm 0.13}$ & $0.72 \pm 0.14$ & $\mathbf{0.84 \pm 0.07}$ & $\mathbf{0.79 \pm 0.11}$\\
    CAM-guided & 0.63 $\pm$ 0.04 & 0.63 $\pm$ 0.15 & 0.83 $\pm$ 0.16 & 0.77 $\pm$ 0.10\\
    Random-guided &0.23\ $\pm 0.18$ & $0.43 \pm 0.40$ & $0.72 \pm 0.20$ & $0.65 \pm 0.27$\\
    Vanilla & $0.21 \pm 0.05$ & $0.59\pm 0.26$ & $0.75 \pm 0.06$ & $0.70 \pm 0.06$\\
    ReasonNet ~\cite{shao2023reasonnet} & $0.63 \pm 0.04$ & $\mathbf{0.73 \pm 0.03}$ & $0.80 \pm 0.04$ & $0.70 \pm 0.06$\\
    \bottomrule
  \end{tabular}
\end{table}

\begin{figure*}[t!]
\centering
\subfloat[The learned human attention and the attention of AEGIS, Vanilla, BC. AEGIS focuses on cars and prioritizes instances ($t2$).]{%
    \includegraphics[width=\textwidth]{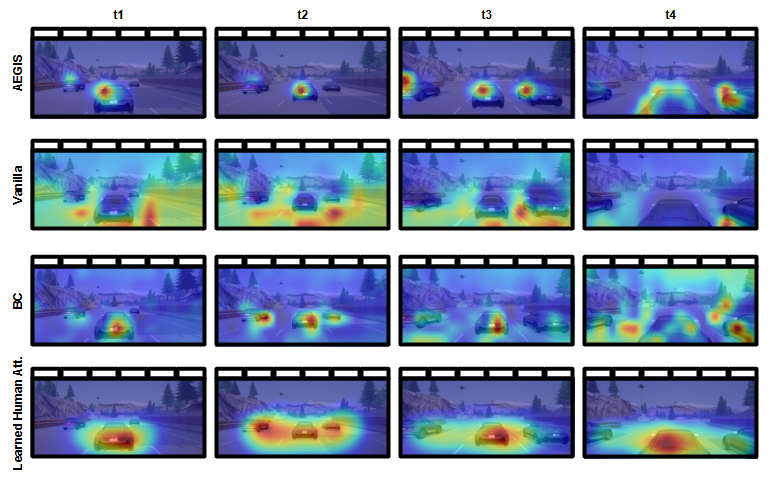}
    \label{fig:fig5a}
}
\vfill
\subfloat[The action of the methods across the time step. The dotted lines show the four time steps of the top figure with the distance to the lead vehicle.]{%
    \includegraphics[width=\textwidth]{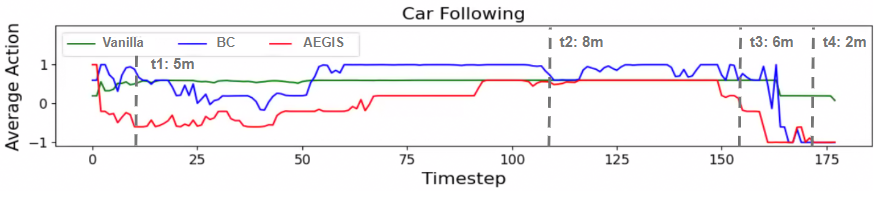}
    \label{fig:fig5b}
}
\caption{The visualization results of attention of the car-following scenario with the fixed action setting, and the average action across time steps for each method.}
\label{fig:fig5}
\end{figure*}

\begin{figure*}[t!]
\centering
\subfloat[The learned human attention and the attention of AEGIS, Vanilla, BC. AEGIS can focus on the critical object.]{%
    \includegraphics[width=\textwidth]{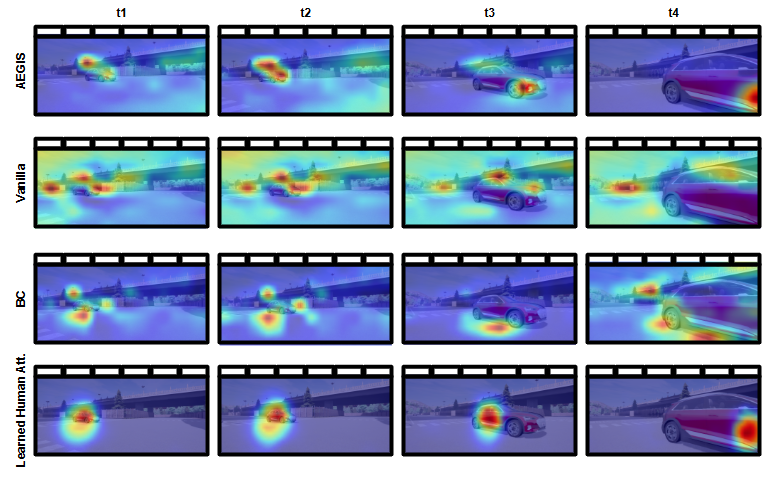}
    \label{fig:fig6a}
}
\vfill
\subfloat[The action of the methods across the time step. The dotted lines show the four time steps of the top figure with the distance to the closest vehicle.]{%
    \includegraphics[width=\textwidth]{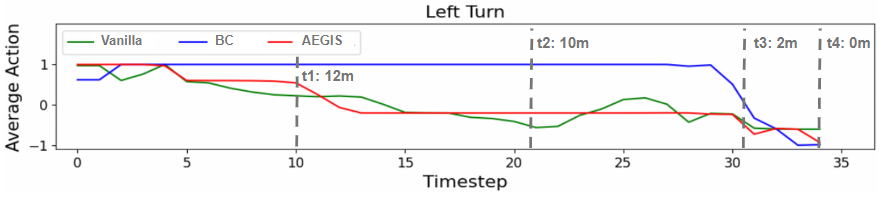}
    \label{fig:fig6b}
}
\caption{The visualization results of attention of the left-turn scenario with the fixed action setting, and the average action across time steps for each method.}
\label{fig:fig6}
\end{figure*}

\subsection{Benchmarks}

In this section, we show the faster convergence speed and training performance of AEGIS in the \textit{Evaluation of Training Phase}, and analyze the better testing performance, generalization ability, decision-making and attention mechanisms of AEGIS in the \textit{Evaluation of Testing Phase}.
\\
\textbf{Evaluation of the training phase.} We first investigate whether learned human attention can improve RL learning by comparing it against Vanilla and BC in our car-following and left-turn scenarios. 
The average reward and survival distance serve as metrics to demonstrate performance across episodes. 
As depicted in Fig. \ref{supp_fig:training_curve}, AEGIS outperforms Vanilla and BC by achieving the highest average reward at the end of training for both scenarios. 
Notably, AEGIS reaches the highest reward achieved by Vanilla in fewer episodes, 270\% faster for the car-following scenario and 150\% faster for the left-turn scenario. 
Note that all methods require similar GPU training time, as the inference time for the pretrained human attention network is merely 0.005 seconds. 
Additionally, AEGIS shows lower variance than Vanilla, indicating more robust performance. 
While BC exhibits a faster convergence speed initially, it does not perform well in the testing scene and thus suffers from severe overfitting issues (see Tab.~\ref{tab:supp_headings} and Tab.~\ref{tab:supp_drop-left-turn}). 
The results of survival distance further verify the statement.
\\
\textbf{Evaluation of testing phase.} We demonstrate that AEGIS outperforms Vanilla and BC in performance with the free action setting on an unseen scene. Tab.~\ref{tab:tab1} shows that AEGIS consistently outperforms the other baselines in two scenarios, achieving an average success rate of 62\% in the car-following scenario and 65\% in the left-turn scenario. 
Moreover, AEGIS achieves the highest survival distance and TTC, showing its superior ability to maintain a safe driving distance and avoid collisions compared to other methods. 
Notably, the success rates of Vanilla and BC decrease significantly by 17\% and 23\% for the car-following scenario and 38\% and 50\% for the left-turn scenario in the new scene, highlighting their poor generalization capabilities, while AEGIS has a minor decrease of 2\% and 6\% for car-following and left-turn scenarios respectively (see Tab.~\ref{tab:supp_headings} and Tab.~\ref{tab:supp_drop-left-turn}).
Although segmentation masks are used as input to mitigate the domain gap, BC and Vanilla still suffer from significant overfitting, whereas AEGIS does not, highlighting the importance of using learned human attention guidance to identify critical objects.

Additionally, we visualize the attention learned by AEGIS, Vanilla, and BC within the car-following and left-turn scenarios, as shown in Fig. \ref{fig:fig5a} and Fig. \ref{fig:fig6a}, respectively. 
In the car-following scenario, AEGIS focuses on the most critical object, the lead vehicles, at all four time steps. 
This aligns with learned human attention (refer to the bottom row of Fig.~\ref{fig:fig5a}), indicating that the AEGIS is effectively guided by learned human attention through our framework. Remarkably, AEGIS is able to prioritize the important instances (see t2 in Fig.~\ref{fig:fig5a}) when the vehicles are far apart and shifts its focus to the surrounding vehicles as they get closer (see t3 in Fig.~\ref{fig:fig5a}). 
In contrast, the machine attention of Vanilla appears more random and is often focused on the ground at all time steps. 
Although the machine attention of BC focuses on the lead vehicle, it is more scattered compared to AEGIS, with unnecessary attention to the background, which is less relevant to the task (see Sec. \ref{sec:human_att_like}).

We further analyze the actions of AEGIS alongside baseline methods at four time steps as presented in Fig.~\ref{fig:fig5a} and Fig.~\ref{fig:fig5b}. 
In the car-following scenario, the lead vehicle is closer to the ego vehicle around t1, moves away around t2, starts braking around t3, and is hit around t4. AEGIS performs well by choosing to initially brake around t1, throttle around t2, brake timely around t3, and ultimately performing a full brake around t4. 
Unlike AEGIS, Vanilla tends to maintain the throttle status most of the time, and BC tends to keep the highest throttle and act quickly. 
In the left-turn scenario, a similar pattern occurs, whereas a difference is that Vanilla acts somewhat akin to AEGIS. 
However, compared with AEGIS, Vanilla RL tends to increase throttle around t2, resulting in less robust and inconsistent decision-making.

\begin{figure*}
  \centering
  \includegraphics[width=0.8\linewidth]{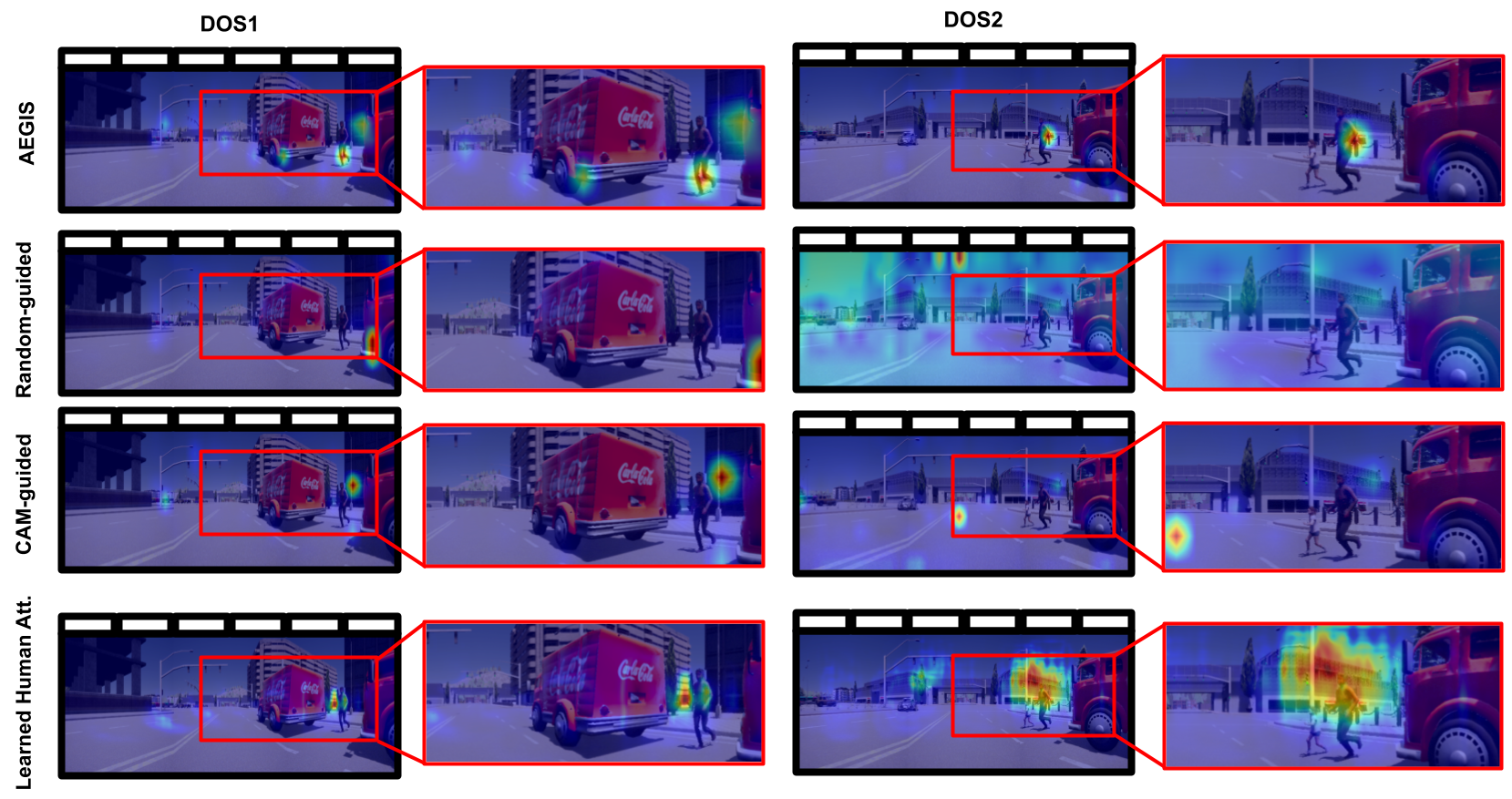}
  \caption{Visualization results of the occlusion scenes with pedestrians (DOS1 and DOS2) with the fixed-action setting. AEGIS successfully identifies the pedestrian near the red vehicle, while other baselines fail to recognize the pedestrian.}
  \label{supp_fig:dos}
\end{figure*}

For benchmark purposes, we further test AEGIS in four diverse occlusion scenarios following~\cite{shao2023reasonnet}. AEGIS has the best overall performance in the DOS benchmark. Remarkably, AEGIS is competitive with ReasonNet~\cite{shao2023reasonnet} in all four occlusion scenarios while maintaining explainability (see Tab.~\ref{tab:tab3}). 
Although the success rate of AEGIS is slightly higher than that of CAM-guided in Tab.~\ref{tab:tab3}, the attention is more reasonable and explainable in Fig.~\ref{supp_fig:dos}.
We also observe that integrating CAM guidance into the framework, even using the basic version~\cite{zhou2016cvpr}, approximately doubles the training time.
The visualization results of machine attention in Fig.~\ref{supp_fig:dos} show that AEGIS effectively identifies pedestrians in scenarios where they suddenly cross the road, allowing the ego vehicle to respond appropriately.

\begin{table*}[htb!]
  \caption{Performance gap between training to evaluation environment in the car-following scenario. AEGIS drops less than Vanilla and BC.
  }
  \label{tab:supp_headings}
  \centering
\begin{tabular}{@{}l@{\hspace{10pt}}c@{\hspace{10pt}}c@{\hspace{10pt}}c}
\toprule
    Model  & Training success rate $\uparrow$ & Testing success rate $\uparrow$  & Drop $\downarrow$ \\
    \midrule
    AEGIS (Ours) &  $0.64 \pm 0.33$ & $\mathbf{0.62 \pm 0.48}$ & $\mathbf{0.02}$ \\
    BC & $\mathbf{0.69 \pm 0.36}$ & $0.46 \pm 0.40$ & $0.23$ \\
    Vanilla & $0.35 \pm 0.43$ &  $0.18 \pm 0.36$ & $0.17$ \\

  \bottomrule
  \end{tabular}
\end{table*}

\begin{table*}[htb!]
  \caption{Performance gap between training to evaluation environment in the left-turn scenario. AEGIS drops less than Vanilla and BC.
  }
  \label{tab:supp_drop-left-turn}
  \centering
\begin{tabular}{@{}l@{\hspace{10pt}}c@{\hspace{10pt}}c@{\hspace{10pt}}c}
\toprule
    Model  & Training success rate $\uparrow$ & Testing success rate $\uparrow$  & Drop $\downarrow$ \\
    \midrule
    AEGIS (Ours) & $ 0.71 \pm 0.21$ & $\mathbf{0.65 \pm 0.27}$ & $\mathbf{0.06}$ \\
    BC & $\mathbf{0.73 \pm 0.16}$ & $0.23 \pm 0.22$ & $0.50$ \\
    Vanilla & $0.71 \pm 0.21$ &  $0.33 \pm 0.36$ & $0.38$ \\
  \bottomrule
  \end{tabular}
\end{table*}

\subsection{Human-like attention and its benefits}
\label{sec:human_att_like}

In this section, we show that the attention of AEGIS is closer to learned human attention in \textit{Similarity between Machine and Learned Human Attention}, focusing more frequently on critical categories (e.g., vehicles or pedestrians) than humans due to the regularization design in \textit{Ratio of Focus Categories}, and exhibiting less scatter in \textit{Spatial Entropy of Machine Attention}, regardless of the focus categories. In addition, we demonstrate that our human attention guidance design is reasonable in \textit{Correlation Between Human Attention and Rewards}, along with quantitative results showcasing improved interpretability in \textit{Interpretability}.
\\
\hypertarget{similarity_target}{\textbf{Similarity between Machine and Learned Human Attention.}}
The machine attention of AEGIS closely aligns with human attention, especially in two \textit{distribution-based} metrics: CC and KL (see Tab. \ref{supp_tab:tab3} and Tab. \ref{supp_tab:tab4}). 
Although the SIM and NSS of AEGIS and BC are similar in the left-turn scenario, the CC and KL of AEGIS in the same scenario demonstrate AEGIS is closer to learned human attention. 
Overall, AEGIS demonstrates greater similarity to learned human attention when considering both scenarios together. Moreover, in the DOS scenario, AEGIS demonstrates greater similarity to learned human attention compared with CAM-guided attention (see Tab.~{\ref{supp_tab:kl_dos}}).
The visualization results in Fig.~\ref{fig:fig5a}, Fig.~\ref{fig:fig6a} and Fig.~\ref{supp_fig:dos} further verify this statement.
\\
\begin{table}[tb]
  \caption{The similarity between human and machine attention in the car-following scenario in an unseen town with a fixed action setting. The attention of AEGIS is more similar to human attention.
  }
  \label{supp_tab:tab3}
  \centering
\begin{tabular}{@{}l@{\hspace{6pt}}c@{\hspace{6pt}}c@{\hspace{6pt}}c@{\hspace{6pt}}c@{}}
\toprule
    Model  & CC $\uparrow$ & KL $\downarrow$  & SIM $\uparrow$ & NSS $\uparrow$\\
    \midrule
    AEGIS (Ours) &  $\mathbf{0.43 \pm 0.03}$ & $\mathbf{2.15 \pm 0.06}$ & $\mathbf{0.62 \pm 0.07}$ & $ \mathbf{0.13 \pm 0.55}$\\
    BC & $0.32 \pm 0.06$ & $2.34 \pm 0.12$ & $0.59 \pm 0.04$ & $-0.23 \pm 1.00$\\
    Vanilla & $0.24 \pm 0.10$ &  $2.47 \pm 0.20$ & $0.52 \pm 0.05$ & $-1.17 \pm 0.89$\\
  \bottomrule
  \end{tabular}
\end{table}
\begin{table}[tb]
  \caption{The similarity between human and machine attention in left-turn scenario in an unseen town with a fixed action setting. The attention of AEGIS is more similar to human attention.
  }
  \label{supp_tab:tab4}
  \centering
\begin{tabular}{@{}l@{\hspace{6pt}}c@{\hspace{6pt}}c@{\hspace{6pt}}c@{\hspace{6pt}}c@{}}
\toprule
    Model  & CC $\uparrow$ & KL $\downarrow$  & SIM $\uparrow$ & NSS $\uparrow$\\
    \midrule
    AEGIS (Ours) & $\mathbf{0.25 \pm 0.08}$ & $\mathbf{4.26 \pm 0.29}$ & $\mathbf{0.34 \pm 0.05}$ & $\mathbf{ 0.20 \pm 0.18}$\\
    BC & $0.21 \pm 0.01$ & $4.42 \pm 0.02$ & $\mathbf{ 0.34 \pm 0.01}$ & $\mathbf{0.20 \pm 0.36}$\\
    Vanilla & $0.18 \pm 0.07$ &  $4.50 \pm 0.14$ & $0.29 \pm 0.07$ & $-0.38 \pm 0.97$\\

  \bottomrule
  \end{tabular}
\end{table}
\begin{table}[tb]
  \caption{KL divergence between machine attention and learned human attention for CAM-guided and AEGIS models across occlusion scenarios (DOS1-DOS4).}
  \label{supp_tab:kl_dos}
  \centering
\begin{tabular}{@{}l@{\hspace{10pt}}c@{\hspace{10pt}}c@{\hspace{10pt}}c@{\hspace{10pt}}c@{}}
\toprule
    Model  & DOS1 & DOS2  & DOS3 & DOS4\\
    \midrule
    AEGIS (Ours) &  $\mathbf{1.86}$ & $\mathbf{1.13}$ & $\mathbf{3.24}$ & $ \mathbf{3.67}$\\
    CAM-guided &  $2.11$ & $1.46$ & $3.74$ & $4.92$\\
  \bottomrule
  \end{tabular}
\end{table}
\begin{figure}[ht]
\centering
\subfloat{%
  \includegraphics[width=\linewidth]{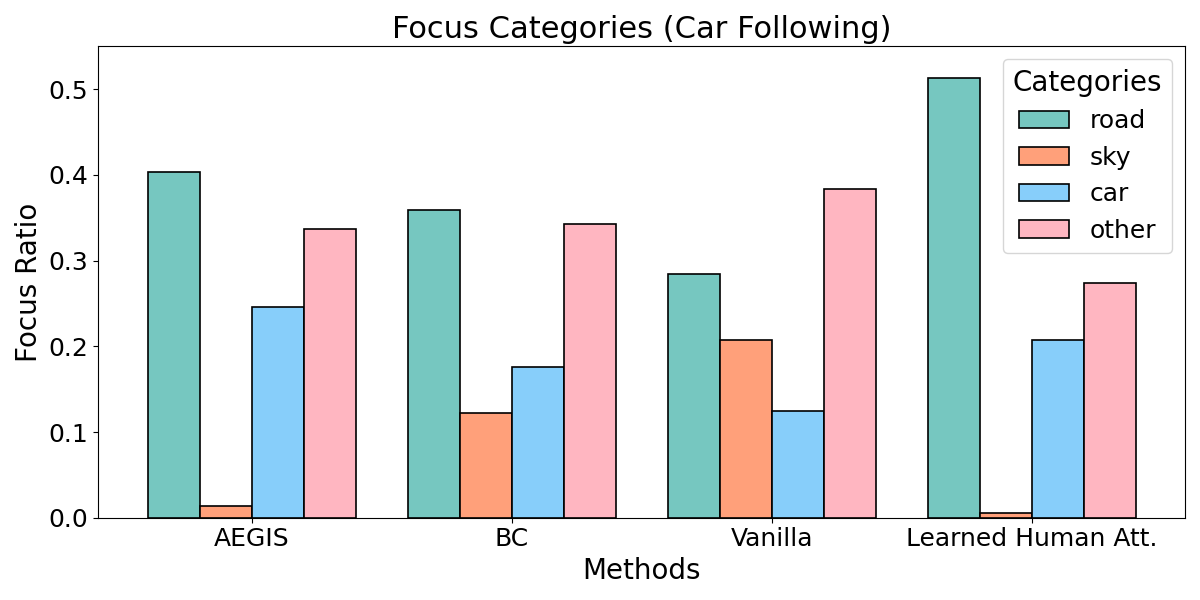}%
  \label{fig:sub-first}
}
\hfill
\subfloat{%
  \includegraphics[width=\linewidth]{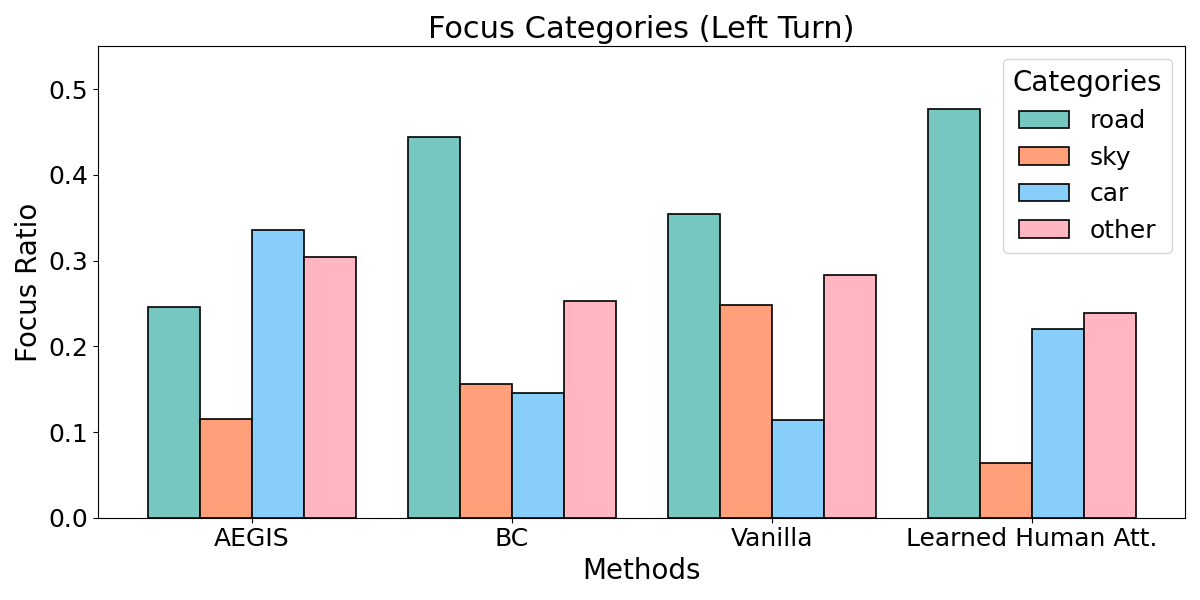}%
  \label{fig:sub-second}
}
\caption{Ratio of focus categories from different methods and learned human attention in the car-following and left-turn scenarios. AEGIS can concentrate on the most crucial object, the car.}
\label{supp_fig:fig_focus}
\end{figure}
\\
\textbf{Ratio of Focus Categories.} We further analyze the focus categories of machine attention from the methods and compare them with human attention (see Fig.~\ref{supp_fig:fig_focus}). 
To precisely quantify the focus categories, we first filter out the less relevant regions. 
This process converts machine attention maps to binary masks using a 0.1 threshold and removes regions with attention levels lower than the threshold. 
Subsequently, we intersect these masks with semantic segmentation data to calculate category-specific ratios. 
Finally, we average these ratios across all the images and models to identify the primary focus areas. 
We find that AEGIS can focus on the most critical objects, the cars, more often than other baseline methods can, with a ratio of 33.5 \% in the left-turn scenario and 24.6\% in the car-following scenario. Although the focus ratio of the road is similar between BC and learned human attention in the left-turn scenario, BC allocates only 14.6\% of its attention to the most critical objects, the cars, whereas learned human attention allocates 22.0\%.
This confirms our design of human attention guidance, which serves as a regularization term and allows the machine to refine the attention on its own further instead of replicating learned human attention identically. For overall similarity between machine attention and learned human attention, please refer to \textit{Similarity between Machine and Learned Human Attention}. 

\begin{figure}[ht]
\centering
\subfloat{%
  \includegraphics[width=0.8\linewidth]{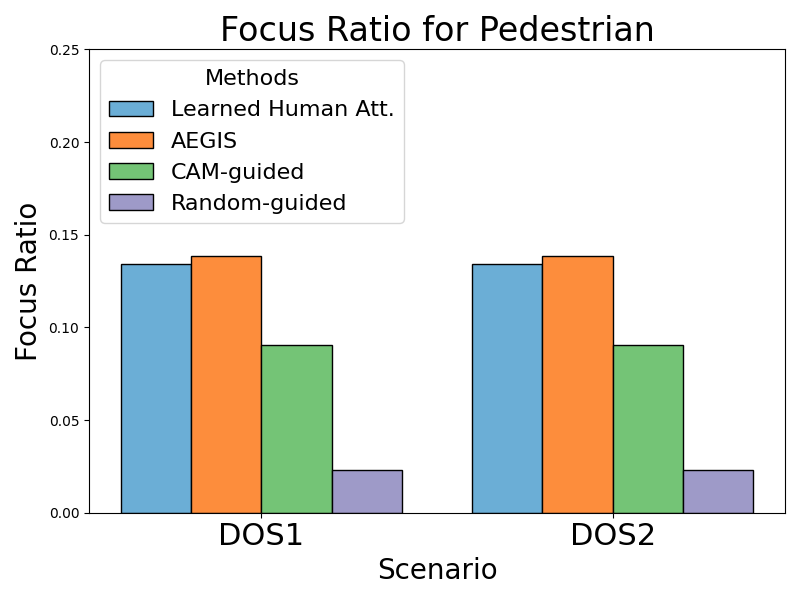}%
  \label{fig:sub-first}
}
\hfill
\subfloat{%
  \includegraphics[width=0.8\linewidth]{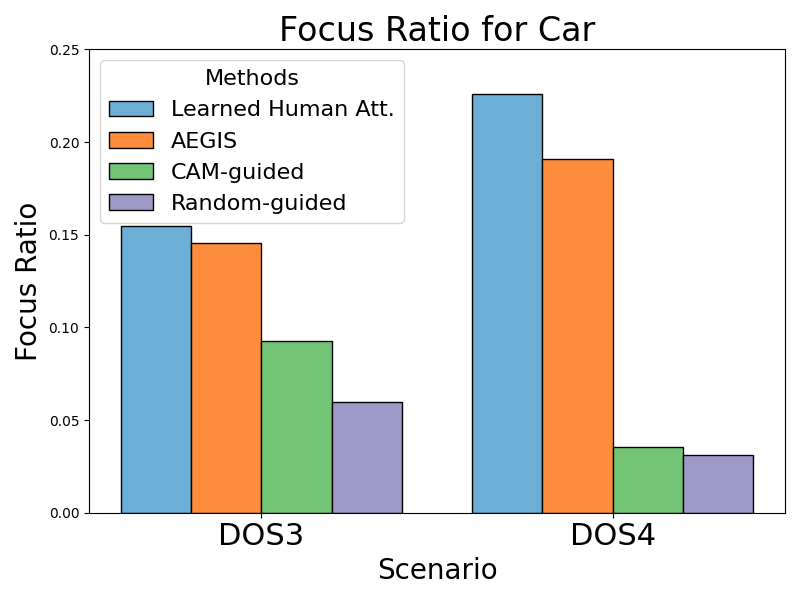}%
  \label{fig:sub-second}
}
\caption{Ratio of focus information for pedestrian-avoidance scenarios (DOS1 and DOS2) and car-avoidance scenarios (DOS3 and DOS4). AEGIS demonstrates a higher focus on dynamic critical objects such as pedestrians in DOS1 and DOS2 and cars in DOS3 and DOS4.}
\label{supp_fig:fig_focus_dos}
\end{figure}

In the DOS3 and DOS4 scenarios (see Fig.~{\ref{supp_fig:fig_focus_dos}}), AEGIS allocates the highest percentage of attention to vehicles compared with other baseline methods, potentially reflecting its prioritization of task-critical elements. In the DOS1 and DOS2 scenarios, which emphasize pedestrian avoidance, AEGIS demonstrates the highest focus on pedestrians.
\\
\label{supp_focus_info_sec}
\begin{table}[htb!]
  \caption{Spatial entropy of the attention maps from car-following and left-turn scenario. The spatial entropy indicates the degree of sparsity in attention maps, regardless of whether the focus is on the critical object or not. The spatial entropy of AEGIS is smaller than Vanilla and BC, indicating more concentrated attention.
  }
  \label{supp_tab:entropy}
  \centering
\begin{tabular}{@{}l@{\hspace{10pt}}c@{\hspace{10pt}}c@{\hspace{10pt}}c}
\toprule
    Model  & Left Turn & Car Following\\
    \midrule
    AEGIS (Ours) &  $ \textbf{0.896}$ & $\textbf{0.869}$ \\
    BC & $0.945$ & $0.928$ \\
    Vanilla & $0.965$ &  $0.954$ \\
    \hline
    Learned human attention & $0.872$ & $0.923$\\
  \bottomrule
  \end{tabular}
\end{table}
\\
\textbf{Spatial Entropy of Machine Attention.} We analyze the spatial entropy of the machine attention to quantify the overall uncertainty of the focus of attention, following the methodology in~\cite{shiferaw2019gaze}. 
We partition the image to a $4\times4$ grid and calculate the spatial entropy \cite{shiferaw2018stationary} via Shannon’s entropy equation \cite{shannon1948mathematical}. 
Notably, spatial entropy illustrates the scattered degree of attention, independent of whether it is directed at critical objects. While BC's spatial entropy is closer to human levels for car following scenario of Tab.~\ref{supp_tab:entropy}, it may not align with human object focus. For instance, in Fig.~\ref{supp_fig:fig_focus}, BC focuses less on the car and more on the sky than learned human attention and AEGIS in the car-following scenario. We prefer smaller spatial entropy than the human attention network, as it indicates that RL agents not only learn human attention patterns but also refine learned human attention by filtering out less relevant objects. This can be observed in Fig.~\ref{fig:fig5a}, where AEGIS in t2 focuses solely on the critical car, while BC in t4 is overly scattered. 

AEGIS has lower spatial entropy than BC and Vanilla in both car-following and left-turn scenarios (see Tab.~\ref{supp_tab:entropy}). 
This illustrates that the machine attention of AEGIS is more concentrated and less random across spatial locations. For overall similarity between machine attention and learned human attention, please refer to \textit{Similarity between Machine and Learned Human Attention}.
\begin{figure}[htb!]
   \centering
   \includegraphics[width=\linewidth]{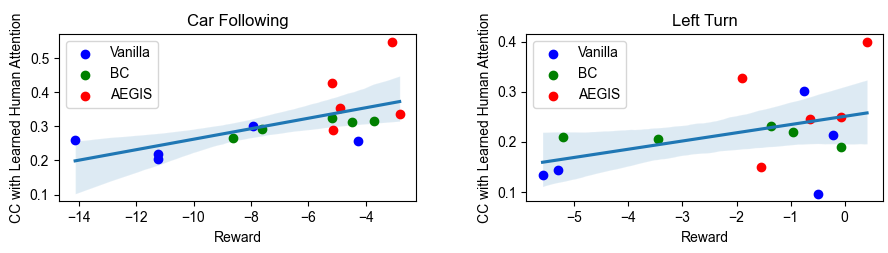}
   \caption{Linear regression of the Pearson's CC between human and machine attention regarding reward. The X-axis represents the reward, and the Y-axis represents the CC (similarity) between machine and learned human attention. The CC is positively correlated with the reward in two scenarios.}
   \label{supp_fig:linear_regression}
\end{figure}
\\
\textbf{Correlation between Human Attention and Rewards.} Like~{\cite{guo2021machine},} we further investigate the relationship between the rewards and the similarity with human attention.
To this end, a linear regression analysis is conducted to examine the correlation between RL rewards and Pearson's correlation coefficient (CC) between human and machine attention (see Fig.~\ref{supp_fig:linear_regression}). The points representing CC and rewards in the linear regression are obtained from the five models trained with different random seeds for each method, as reported in Tab.~\ref{supp_tab:tab3} and Tab.~\ref{supp_tab:tab4}. The X-axis represents the reward, and the Y-axis shows the CC (similarity) between machine and learned human attention. Notably, AEGIS has the highest average CC (see red dots in Fig.~\ref{supp_fig:linear_regression}, Tab.~\ref{supp_tab:tab3} and Tab.~\ref{supp_tab:tab4}). In both scenarios, rewards (performance) are positively correlated with CC between machine and learned human attention. In the left-turn scenario, the linear regression model yields an R-squared value of 0.377 $(p = 0.0148)$. These results indicate that the positive correlation between CC and RL rewards is statistically significant at conventional significance levels. For the car-following scenario, while a positive correlation is observed with an R-squared value of 0.128, the associated \textit{p} value of 0.19 indicates that this correlation does not achieve statistical significance.
\\
\textbf{Interpretability.}
In this work, interpretability is defined as the degree to which a model's decision-making process can be understood and explained by humans~\cite{ali2023explainable}.
Naturally, if a model's machine attention map is closer to human attention, the model becomes more interpretable and human-understandable. 
As shown in Tab.~\ref{supp_tab:tab3} and Tab.~\ref{supp_tab:tab4}, AEGIS achieves closer alignment with human attention. We also conducted a survey with 80 human participants, asking three questions related to the interpretability and safety of the model. The participants were compensated at a rate higher than the country's law. The survey involved showing videos of AEGIS, Vanilla, and BC in the video figure.

\begin{itemize}
    \item \textbf{QA}: Rank the three videos from easiest to most difficult to understand, based on the relationship between the visualization and the action.
    \item \textbf{QB}: Rank the three videos based on how well their visualizations and actions meet your expectations.
    \item \textbf{QC}: Rank the three videos based on how confident you feel about their safety and reliability in performing autonomous driving tasks.
\end{itemize}

\begin{figure}[ht]
\centering
\subfloat{
    \includegraphics[width=\linewidth]{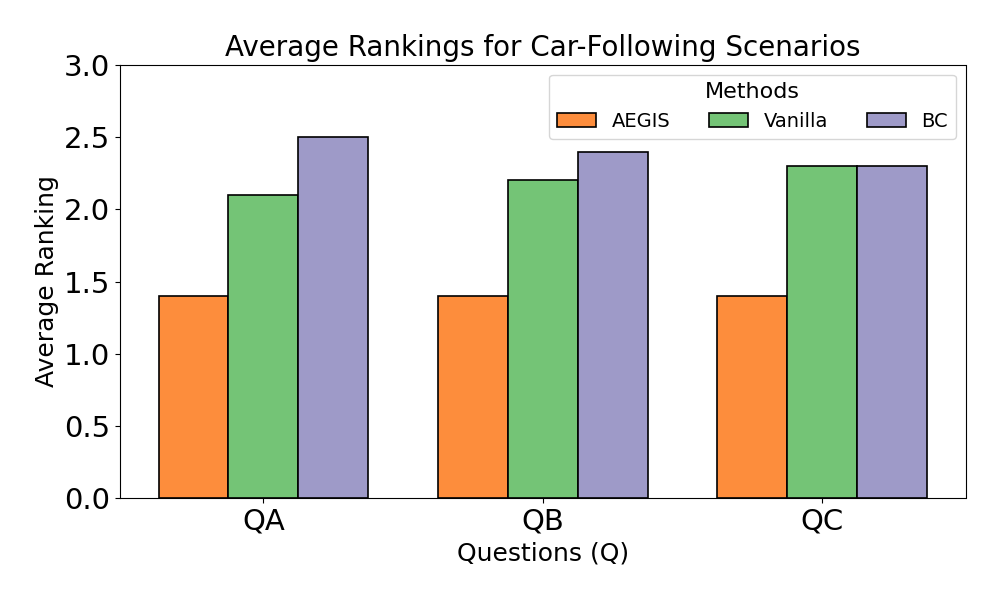}%
}
\hfill
\subfloat{
    \includegraphics[width=\linewidth]{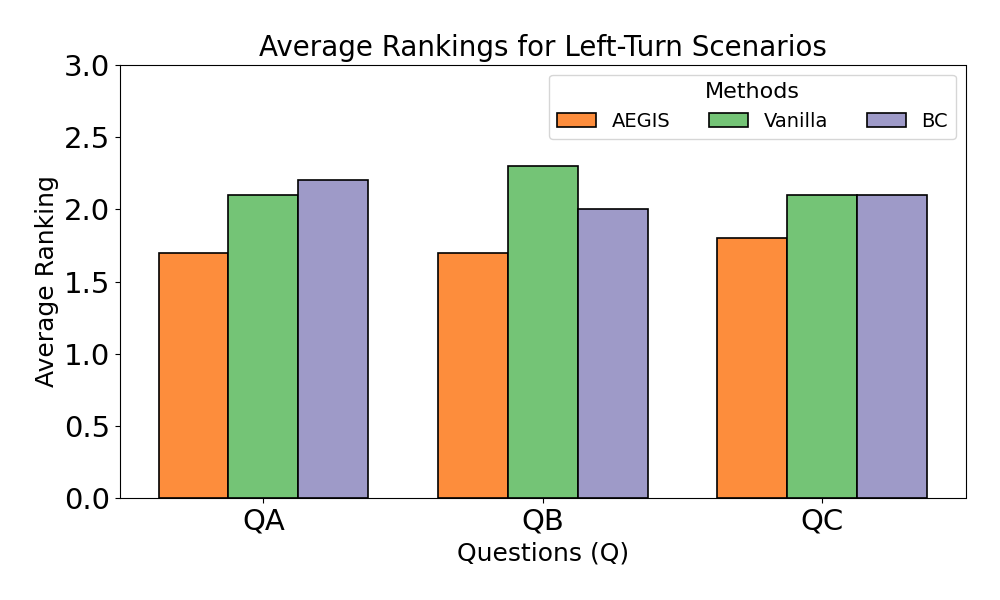}%
}
\caption{The average rankings for interpretability (QA and QB) and safety (QC), derived from a survey conducted with 80 participants, indicate that AEGIS improves both interpretability and safety.}
\label{fig:q_human}
\end{figure}

QA and QB focus on the interpretability of the models, aiming to identify which model is more human-understandable, while QC evaluates the safety and reliability of the models. 
Fig.~{\ref{fig:q_human}} demonstrates that AEGIS achieves better average rankings, indicating that human attention guidance can enhance both interpretability and safety.

\section{Discussion}
\begin{figure*}[htb!]
  \centering
  \includegraphics[width=0.8\linewidth]{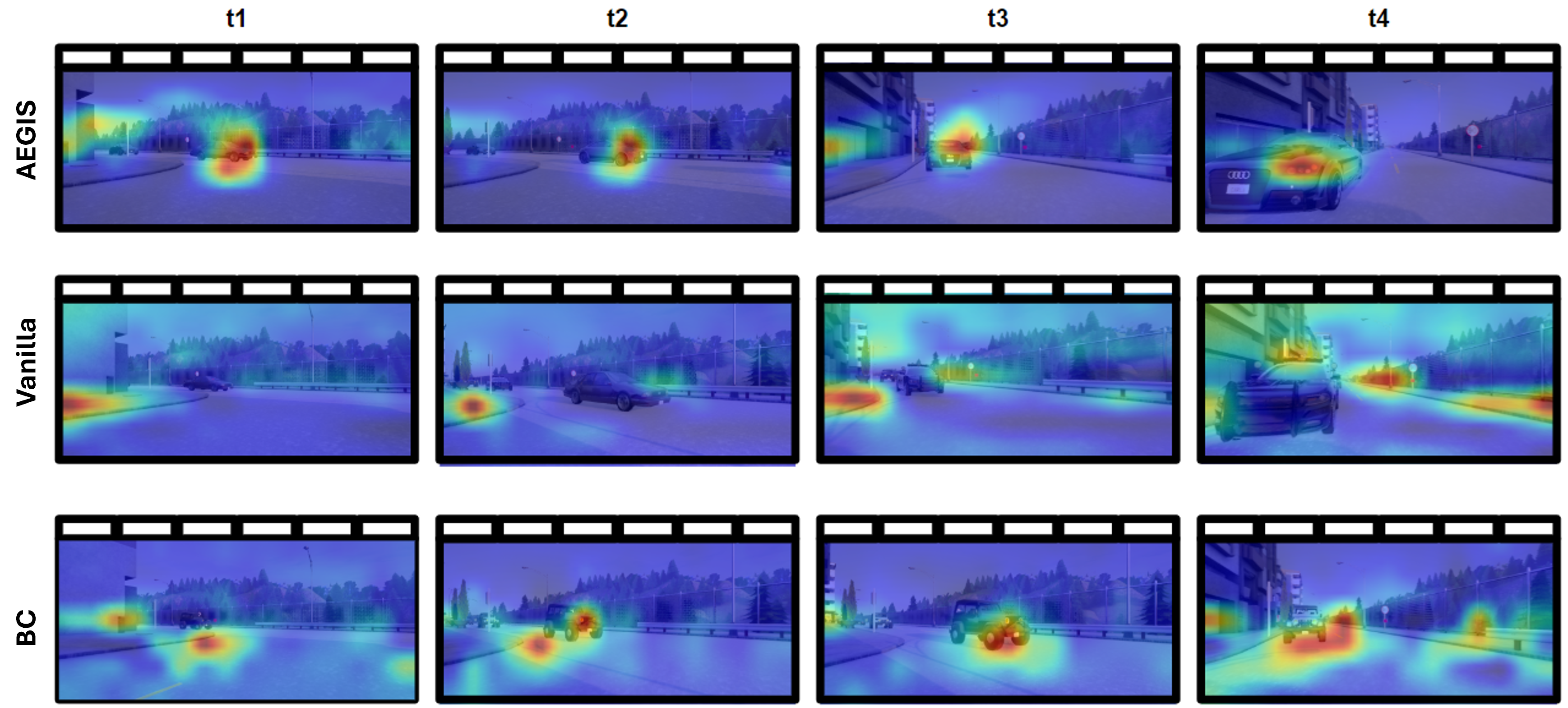}
  \caption{Visualization results of the left-turn scene with the free-action setting during training. Vanilla and BC overfit on the non-vehicle objects, leading to the performance drop during testing.}
  \label{supp_fig:left-turn-train}
\end{figure*}

In this work, we present AEGIS, a framework to increase the interpretability and performance of intelligent vehicles via human attention guidance. 
By aligning machine attention with learned human attention closely in the early training phase of RL, AEGIS can learn to focus on task-relevant objects. 
This is essential for preventing overfitting in RL training, as demonstrated in Fig.~\ref{supp_fig:left-turn-train}. 
Both the Vanilla RL and AEGIS can successfully execute a left turn. 
However, AEGIS performs the maneuver based on the correct cue, the oncoming vehicle, whereas the Vanilla RL incorrectly bases its action on the sidewalk.
This is supported by the dramatic drop of 38\% in the performance of Vanilla from training to testing (see Tab.~\ref{tab:supp_headings} and Tab.~\ref{tab:supp_drop-left-turn}). 
Our framework can improve the interpretability of current RL, evidenced by higher similarity with learned human attention and higher ranking in the interpretability survey.
In addition, we propose the largest human eye-tracking dataset that is designed for the task.
Compared with existing in-car datasets,
Our dataset faces fewer ethical concerns, has lower costs, and can provide repeatable scenarios for different drivers. 
Compared with other in-lab datasets, our dataset offers a more immersive experience using a VR headset.


\subsection{Human attention network}
\begin{figure*}[htb!]
  \centering
  \includegraphics[width=0.8\linewidth]{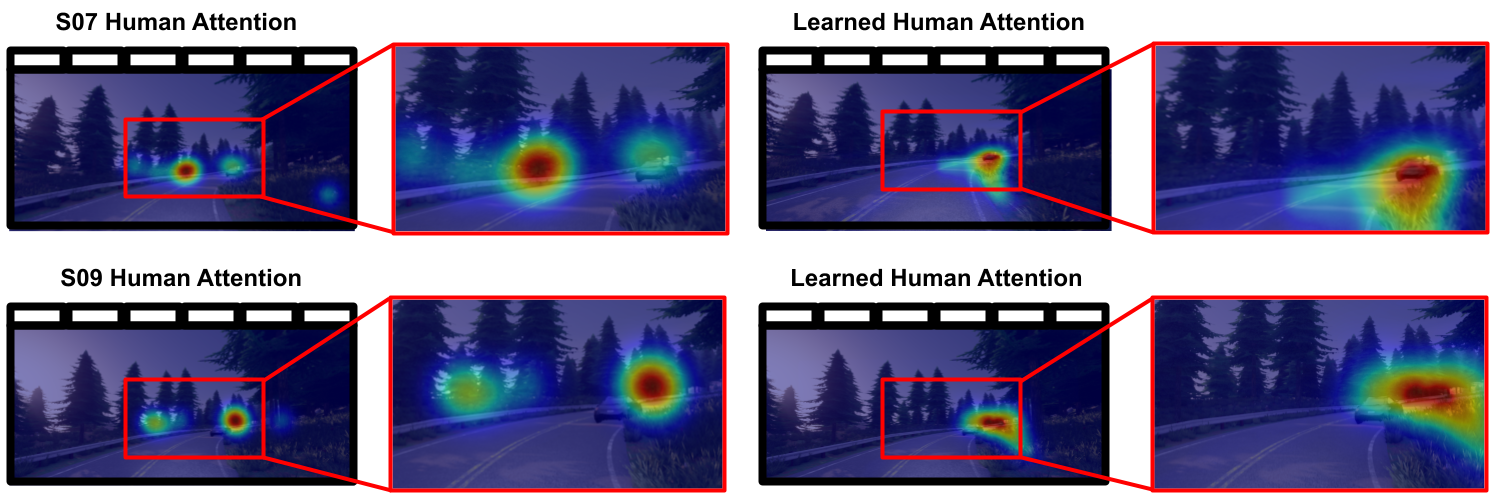}
  \caption{Human attention from subjects S07 and S09 reveals different focus patterns in similar scenes. S07's attention is primarily directed toward the road fence, while S09 focuses more on the lead vehicle. Incorporating a human attention model can help mitigate distraction issues.}
  \label{fig:supp_eye_tracking}
\end{figure*}

The pre-trained human attention network accurately predicts where most people are likely to focus their attention in a scene. For example, the human attention network can identify the lead vehicle in a car-following scenario (see Fig. \ref{fig:supp_eye_tracking}).
This network ensures consistency in predicting learned human attention across participants, who may exhibit diverse eye-tracking patterns. For example, in Fig.~\ref{fig:supp_eye_tracking}, S07’s attention is primarily directed toward the road fence, whereas S09 focuses more on the lead vehicle, however, the learned human attention can predict where most people look. Additionally, the human attention network enables the prediction of human attention across different observations, allowing agents to take their actions. The use of a pre-trained network also reduces the burden on human operators in practical applications, as the pre-trained human attention network eliminates the need for human involvement during RL training unlike human-in-the-loop RL in which humans need to stand by~\cite{wu2023human, wu2022prioritized, wu2023toward}, and it allows human-free inference which can have broader application when human gaze data are unavailable. 

To evaluate the central bias of the human attention model, we adopt the approach from \cite{palazzi2018predicting, deng_CDNN} and employ information gain (IG) \cite{kummerer2015information} as a metric. IG measures the quality of the learned human attention model by comparing learned human attention to ground truth human attention and a baseline map. For this analysis, we use a centered Gaussian baseline \cite{deng_CDNN} as the baseline map. An IG score greater than zero indicates that the learned human attention surpasses the centered Gaussian baseline in predicting human attention, thereby demonstrating less central bias. The learned human attention achieves a better KL, NSS, SIM, CC than the centered Gaussian baseline. An IG score much greater than zero not only demonstrates reduced central bias but also suggests the network's capability to predict task-driven changes in gaze direction.

\begin{table}[h!]
\caption{KL, NSS, SIM, CC, and Information Gain (IG) are used to evaluate the learned human attention compared to a centered Gaussian. Lower KL value and higher NSS, SIM, CC, and IG indicate that the learned human attention is not simply predicting attention at the center of the image, demonstrating its effectiveness beyond a mean predictor.}
\centering
\begin{tabular}{@{}l@{\hspace{10pt}}c@{\hspace{10pt}}c@{\hspace{10pt}}c@{\hspace{10pt}}c@{\hspace{10pt}}c@{}}
    \toprule
Attention & KL$\downarrow$ & NSS $\uparrow$ & SIM $\uparrow$ & CC $\uparrow$ & IG $\uparrow$ \\ \midrule
learned human attention & \textbf{2.11} & \textbf{0.54} & \textbf{0.37} & \textbf{0.46} & \textbf{4.95} \\
centered Gaussian & 6.27 & 0.52 & 0.13 & 0.16 & 0.00 \\ \bottomrule
\end{tabular}
\label{tab:model_results}
\end{table}

\label{sec:human_att_network}

\subsection{Analysis of tolerance to noisy input}

\begin{figure}
  \centering
  \includegraphics[width=\linewidth]{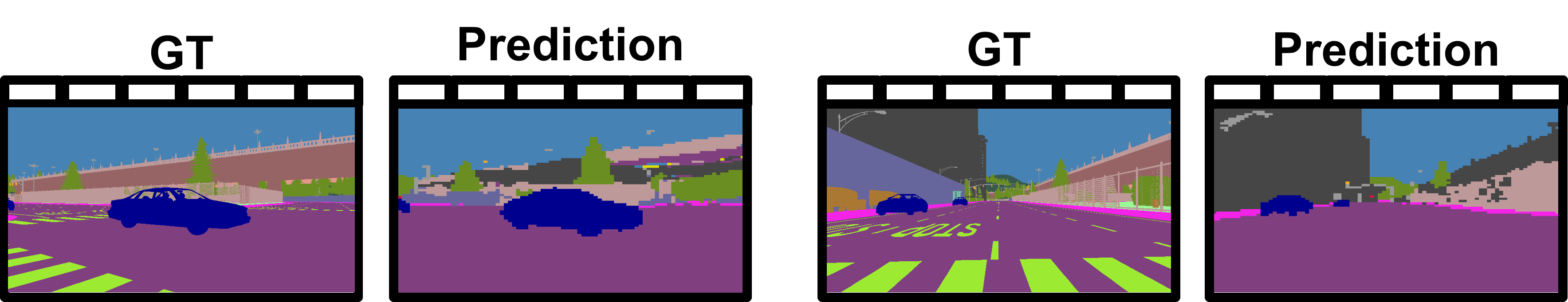}
  \caption{Comparision between segmentation images directly from the simulator (GT) and segmentation images from pre-trained Segformer (Prediction).}
  \label{supp_fig:segformer}
\end{figure}

In this study, the RL model is fed with perfect segmentation images from the simulator. 
To assess the model's robustness and reliability, we investigate the resilience of our model to noise by using RGB images as the input and employing a pre-trained segmentation model to obtain segmentation images. 
We employ Segformer \cite{xie2021segformer} pre-trained on the Cityscapes \cite{cordts2016cityscapes} dataset, resulting in noisier segmentation images than the segmentation directly from the simulator (see Fig.~\ref{supp_fig:segformer}). 
Note that the segmentation images from the CARLA simulator contain customized classes like bridges and road lines that are not presented in Cityscapes. 
We then evaluate the trained RL agents in the noisier segmentation images. AEGIS outperforms BC and Vanilla with a success rate of 55\% and 57\% in the left-turn and car-following scenarios, respectively (see Tab.~\ref{supp_tab:segformer}). 
This analysis reveals the potential of our model to maintain high performance even in less-than-ideal conditions.
\label{supp_sec:seg_analysis}

\begin{table}
  \caption{Evaluation results of the success rate using segmentation images from Segformer. AEGIS outperforms BC and Vanilla.
  }
  \label{supp_tab:segformer}
  \centering
\begin{tabular}{@{}l@{\hspace{6pt}}c@{\hspace{6pt}}c@{\hspace{6pt}}c}
\toprule
    Model  & Left turn & Car following  \\
    \midrule
    AEGIS (Ours) & \bf{ $0.55 \pm 0.14$} & $0.57 \pm 0.31$  \\
    BC & $0.26 \pm 0.14$ &  $0.17 \pm 0.20$ \\
    Vanilla & $0.14 \pm 0.26$ & $0.17 \pm 0.34$  \\

  \bottomrule
  \end{tabular}
\end{table}

\subsection{Impact of input frame number}

For all the experiments, we use three frames of images as the input of RL. Tab.~\ref{supp_tab:frame_size}, we investigate the impact of temporal information by varying the number of input frames. 
Our finding suggests that using 3 frames can achieve the best results in both scenarios. 
The 1-frame setting does not include sufficient temporal information and, thereby, cannot accurately predict the vehicle dynamics. 
On the other hand, the 5-frame setting also degrades performance, likely due to the model's limited capacity to process the increased information effectively. 
Therefore, we adopt the 3-frame setting for all the analyses to achieve a balance between incorporating sufficient temporal information and maintaining manageable model capacity.
\begin{table}[htb!]
  \caption{Performance of AEGIS when training with 1, 3, 5 frames. 3-frames has the best performance due to a balance of model capacity and temporal information.
  }
  \label{supp_tab:frame_size}
  \centering
\begin{tabular}{@{}l@{\hspace{10pt}}c@{\hspace{10pt}}c@{\hspace{10pt}}c@{\hspace{10pt}}c}
\toprule
    Scenario  & 1 frame & 3 frame  & 5 frame \\
    \midrule
   left turn& $0.32 \pm 0.10$  & $0.65 \pm 0.27$ & $0.27 \pm 0.12$ \\
   car following & $0.14 \pm 0.28$ & $0.62 \pm 0.48$ & $0.43 \pm 0.47$ \\

  \bottomrule
  \end{tabular}
\end{table}
\subsection{Analysis of failure cases}
\label{supp_sec:fail_cases}
\begin{figure}[h!]
  \centering
  \includegraphics[width=\linewidth]{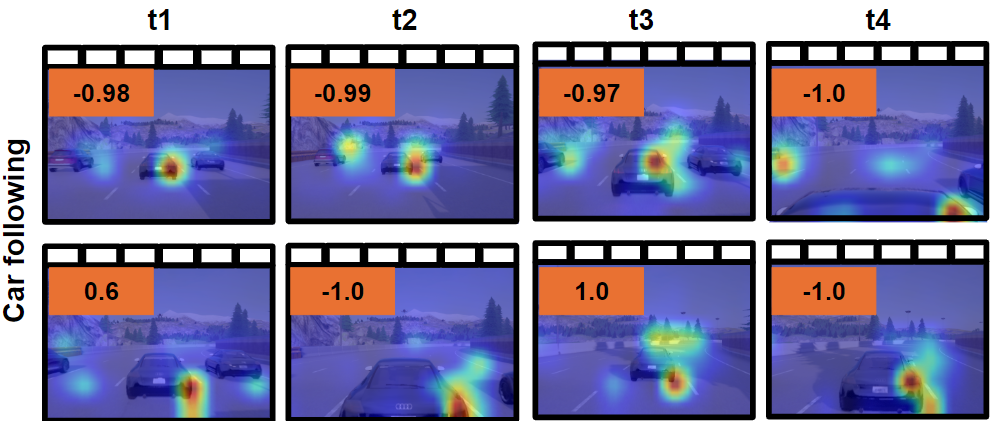}
  \caption{Failure cases of AEGIS with action value at the top-left side in the car-following scene. In the top row, the vehicle collides with the lead vehicle due to failure to maintain a safe following distance. In the bottom row, the vehicle collides with the front car in the road curve (t3-t4).}
  \label{supp_fig:car_following_fail}
\end{figure}

\begin{figure}[htb!]
  \centering
  \includegraphics[width=\linewidth]{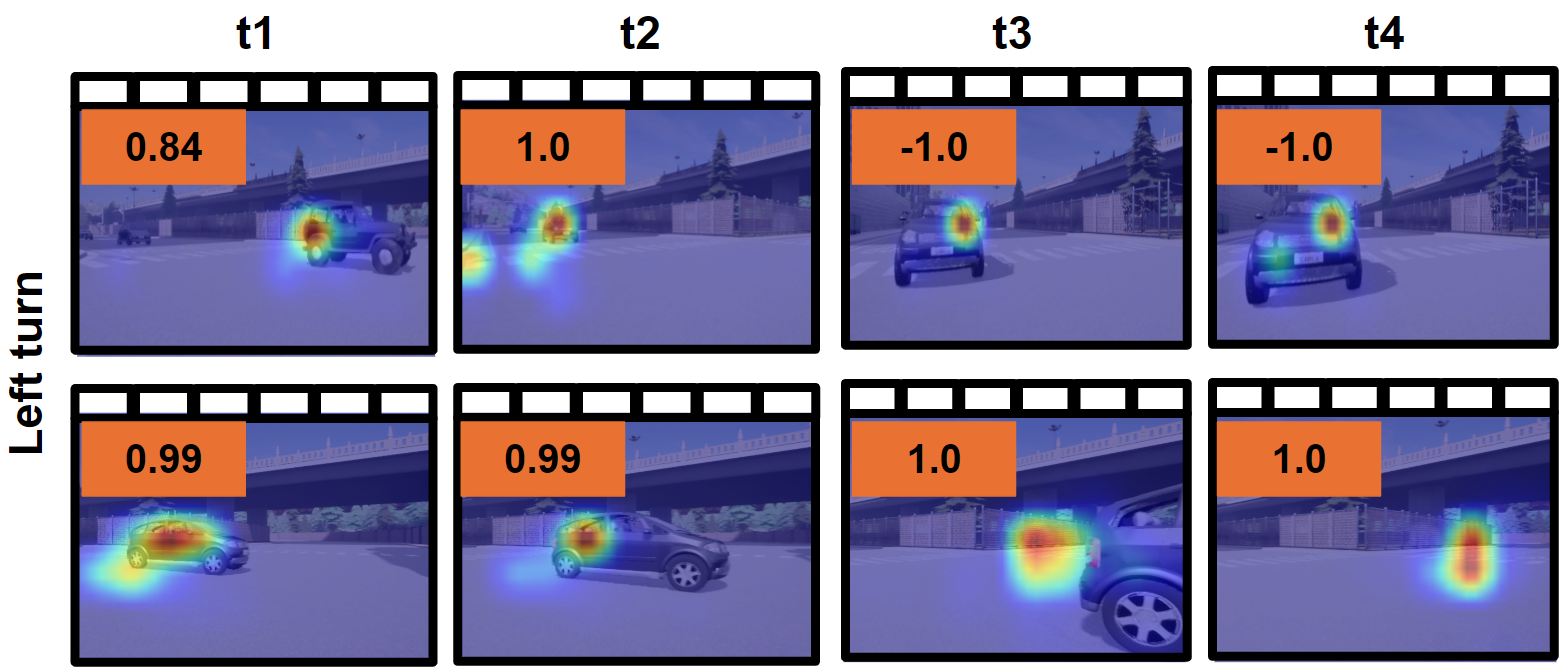}
  \caption{Failure cases of AEGIS with action value at the top-left side in the left-turn scene. In the top row, the vehicle chooses to stop as it cannot safely cross the junction. In the bottom row, the rear of the vehicle collides with another vehicle that has just passed through.}
  \label{supp_fig:left_turn_fail}
\end{figure}

We collect a few interesting failure cases from AEGIS. 
In the top row of Fig.~\ref{supp_fig:car_following_fail}, the vehicle is unable to maintain a safe distance and collides with the car ahead despite executing an immediate braking decision. 
In the bottom row of Fig.~\ref{supp_fig:car_following_fail}, the vehicle successfully avoids the collision on the straight road (t1 and t2), while hitting the lead vehicle on the curved road (t3 and t4). 
The visualization results suggest that the vehicle's attention remains fixated on the curved corner, which may mean it is anticipating incoming traffic. 
However, this may hinder its ability to clearly see the lead vehicle and end up in a collision. 
In the top row of Fig.~\ref{supp_fig:left_turn_fail}, the vehicle detects an approaching vehicle moving quickly towards it, leading it to choose to stop. 
This reaction closely mirrors human behavior, as humans tend to stop the vehicle in emergency situations. 
In the bottom row of Fig.~\ref{supp_fig:left_turn_fail}, as the ego vehicle accelerates during a left turn, it collides with another vehicle almost completely past the intersection. 
This collision happens because the ego vehicle incorrectly estimates that there is sufficient space to complete its turn, failing to account for the other vehicle's proximity and speed.

\subsection{Limitation and future work}
\label{supp_sec:limit}

Although AEGIS shows improved success rates, quicker training speeds, and robust performance across different scenes, our current focus is limited to collision avoidance, with an emphasis on the vehicle's speed. Furthermore, the RL agent in the current work still relies on hand-crafted reward functions, which requires hyperparameter tuning.
Our goal is to develop a framework that leverages human attention to guide the RL agent and demonstrate that the RL agent can benefit from human attention rather than presenting a state-of-the-art method and claiming that attention is all you need.
In future research, we aim to gather trajectory data from human drivers in more scenarios. 
However, collecting long-term and diverse driving data has been challenging due to motion sickness experienced by participants using VR headsets. 
Consequently, we intend to acquire more extensive and varied data from participants without motion sickness in future studies.
Our works shed light on developing explainable guidance for RL autonomous driving tasks using human data.
Although we do not claim that "attention is all you need" and that AEGIS can generalize well to any unseen scene, one interesting future step is to continue maintaining and augmenting the dataset with more scenarios.
Another future step is to collect more human action data for steering control to create generalizable and scalable RL models.

Currently, this work is limited to simulations, as the simulator provides a cost-effective environment for closed-loop training compared to real-world data collection. To explore the potential real-world applicability of our model, we conducted a case study visualizing the learned machine attention on a single real-world video, as shown in the video figure. While the results demonstrate promise, further investigation and testing across a broader range of real-world scenarios are necessary to assess and improve the model's generalizability in future work.
\section{Conclusion}
\label{sec:conclusion}

This paper presents a novel human attention-based explainable guidance for intelligent vehicle systems (AEGIS) framework as a solution to enhance the RL agent's learning efficiency, generalization, and interpretability in complex driving scenarios. 
We collect {eye-movement} data from a VR simulator with 20 participants through our in-lab realistic data collection system and design a policy network with the self-attention layer guided by learned human attention. 
With human attention guidance, AEGIS achieves the highest reward in a shorter timeframe in the car-following and left-turn scenarios. Moreover, AEGIS maintains the highest success rate in unseen maps of both scenarios, highlighting its robustness and generalization capability.
We also conduct further analysis on the focus information of machine attention learned by AEGIS. It demonstrates that AEGIS can learn to prioritize task-relevant objects more effectively by aligning machine attention more closely with human attention. Moreover, a survey with 80 participants demonstrates the attention and action of AEGIS is more interpretable compared with methods without human attention guidance. The study implies the potential of incorporating human attention in the development of AIVs and emphasizes the benefits of integrating human cognitive processes with machine learning algorithms in autonomous driving.

\begin{acks}
This work was partly supported by the Australian Research Council (ARC) under discovery grants DP210101093 and DP220100803 and the UTS Human-Centric AI Centre funding sponsored by GrapheneX (2023-2031). Research was partially sponsored by the Australia Defence Innovation Hub under Contract No. P18650825, Australian Cooperative Research Centres Projects (CRC-P) Round11 CRCPXI000007, USOfficeofNavalResearchGlobal under Cooperative Agreement Number ONRG- NICOP- N62909-19-12058, and AFOSR– DST Australian Autonomy Initiative agreement ID10134. We also thank the NSW Defence Innovation Network and the NSW State Government of Australia for financial support in part of this research through grant DINPP2019 S1-03/09 and PP2122.03.02. We also thank Yi-Shan Hung for contribution in the figures. Special thanks to Mrs Haiting Lan for proofreading.
\end{acks}

\bibliographystyle{ACM-Reference-Format}
\bibliography{egbib}


\begin{thebibliography}{97}


\ifx \showCODEN    \undefined \def \showCODEN     #1{\unskip}     \fi
\ifx \showDOI      \undefined \def \showDOI       #1{#1}\fi
\ifx \showISBNx    \undefined \def \showISBNx     #1{\unskip}     \fi
\ifx \showISBNxiii \undefined \def \showISBNxiii  #1{\unskip}     \fi
\ifx \showISSN     \undefined \def \showISSN      #1{\unskip}     \fi
\ifx \showLCCN     \undefined \def \showLCCN      #1{\unskip}     \fi
\ifx \shownote     \undefined \def \shownote      #1{#1}          \fi
\ifx \showarticletitle \undefined \def \showarticletitle #1{#1}   \fi
\ifx \showURL      \undefined \def \showURL       {\relax}        \fi
\providecommand\bibfield[2]{#2}
\providecommand\bibinfo[2]{#2}
\providecommand\natexlab[1]{#1}
\providecommand\showeprint[2][]{arXiv:#2}

\bibitem[AIA(2024)]%
        {AIAct2024}
 \bibinfo{year}{2024}\natexlab{}.
\newblock \bibinfo{title}{Artificial Intelligence (AI) Act: Council Gives Final
  Green Light to the First Worldwide Rules on AI}.
\newblock
  \bibinfo{howpublished}{\url{https://www.consilium.europa.eu/en/press/press-releases/2024/05/21/artificial-intelligence-ai-act-council-gives-final-green-light-to-the-first-worldwide-rules-on-ai/}}.
\newblock
\newblock
\shownote{Accessed: 2024-11-27}.


\bibitem[Adadi and Berrada(2018)]%
        {adadi2018peeking}
\bibfield{author}{\bibinfo{person}{Amina Adadi} {and} \bibinfo{person}{Mohammed
  Berrada}.} \bibinfo{year}{2018}\natexlab{}.
\newblock \showarticletitle{Peeking inside the black-box: a survey on
  explainable artificial intelligence (XAI)}.
\newblock \bibinfo{journal}{\emph{IEEE access}}  \bibinfo{volume}{6}
  (\bibinfo{year}{2018}), \bibinfo{pages}{52138--52160}.
\newblock


\bibitem[Ali et~al\mbox{.}(2023)]%
        {ali2023explainable}
\bibfield{author}{\bibinfo{person}{Sajid Ali}, \bibinfo{person}{Tamer Abuhmed},
  \bibinfo{person}{Shaker El-Sappagh}, \bibinfo{person}{Khan Muhammad},
  \bibinfo{person}{Jose~M Alonso-Moral}, \bibinfo{person}{Roberto
  Confalonieri}, \bibinfo{person}{Riccardo Guidotti}, \bibinfo{person}{Javier
  Del~Ser}, \bibinfo{person}{Natalia D{\'\i}az-Rodr{\'\i}guez}, {and}
  \bibinfo{person}{Francisco Herrera}.} \bibinfo{year}{2023}\natexlab{}.
\newblock \showarticletitle{Explainable Artificial Intelligence (XAI): What we
  know and what is left to attain Trustworthy Artificial Intelligence}.
\newblock \bibinfo{journal}{\emph{Information fusion}}  \bibinfo{volume}{99}
  (\bibinfo{year}{2023}), \bibinfo{pages}{101805}.
\newblock


\bibitem[Alletto et~al\mbox{.}(2016)]%
        {alletto2016dr}
\bibfield{author}{\bibinfo{person}{Stefano Alletto}, \bibinfo{person}{Andrea
  Palazzi}, \bibinfo{person}{Francesco Solera}, \bibinfo{person}{Simone
  Calderara}, {and} \bibinfo{person}{Rita Cucchiara}.}
  \bibinfo{year}{2016}\natexlab{}.
\newblock \showarticletitle{Dr (eye) ve: a dataset for attention-based tasks
  with applications to autonomous and assisted driving}. In
  \bibinfo{booktitle}{\emph{Proceedings of the ieee conference on computer
  vision and pattern recognition workshops}}. \bibinfo{pages}{54--60}.
\newblock


\bibitem[Arrieta et~al\mbox{.}(2020)]%
        {arrieta2020explainable}
\bibfield{author}{\bibinfo{person}{Alejandro~Barredo Arrieta},
  \bibinfo{person}{Natalia D{\'\i}az-Rodr{\'\i}guez}, \bibinfo{person}{Javier
  Del~Ser}, \bibinfo{person}{Adrien Bennetot}, \bibinfo{person}{Siham Tabik},
  \bibinfo{person}{Alberto Barbado}, \bibinfo{person}{Salvador Garc{\'\i}a},
  \bibinfo{person}{Sergio Gil-L{\'o}pez}, \bibinfo{person}{Daniel Molina},
  \bibinfo{person}{Richard Benjamins}, {et~al\mbox{.}}}
  \bibinfo{year}{2020}\natexlab{}.
\newblock \showarticletitle{Explainable Artificial Intelligence (XAI):
  Concepts, taxonomies, opportunities and challenges toward responsible AI}.
\newblock \bibinfo{journal}{\emph{Information fusion}}  \bibinfo{volume}{58}
  (\bibinfo{year}{2020}), \bibinfo{pages}{82--115}.
\newblock


\bibitem[Atakishiyev et~al\mbox{.}(2024)]%
        {atakishiyev2024explainable}
\bibfield{author}{\bibinfo{person}{Shahin Atakishiyev},
  \bibinfo{person}{Mohammad Salameh}, \bibinfo{person}{Hengshuai Yao}, {and}
  \bibinfo{person}{Randy Goebel}.} \bibinfo{year}{2024}\natexlab{}.
\newblock \showarticletitle{Explainable artificial intelligence for autonomous
  driving: A comprehensive overview and field guide for future research
  directions}.
\newblock \bibinfo{journal}{\emph{IEEE Access}} (\bibinfo{year}{2024}).
\newblock


\bibitem[Baee et~al\mbox{.}(2021)]%
        {Baee_2021_ICCV}
\bibfield{author}{\bibinfo{person}{Sonia Baee}, \bibinfo{person}{Erfan
  Pakdamanian}, \bibinfo{person}{Inki Kim}, \bibinfo{person}{Lu Feng},
  \bibinfo{person}{Vicente Ordonez}, {and} \bibinfo{person}{Laura Barnes}.}
  \bibinfo{year}{2021}\natexlab{}.
\newblock \showarticletitle{MEDIRL: Predicting the Visual Attention of Drivers
  via Maximum Entropy Deep Inverse Reinforcement Learning}. In
  \bibinfo{booktitle}{\emph{Proceedings of the IEEE/CVF International
  Conference on Computer Vision (ICCV)}}. \bibinfo{pages}{13178--13188}.
\newblock


\bibitem[Borji et~al\mbox{.}(2011)]%
        {borji2011computational}
\bibfield{author}{\bibinfo{person}{Ali Borji}, \bibinfo{person}{Dicky~N
  Sihite}, {and} \bibinfo{person}{Laurent Itti}.}
  \bibinfo{year}{2011}\natexlab{}.
\newblock \showarticletitle{Computational modeling of top-down visual attention
  in interactive environments.}. In \bibinfo{booktitle}{\emph{BMVC}},
  Vol.~\bibinfo{volume}{85}. \bibinfo{pages}{1--12}.
\newblock


\bibitem[Bylinskii et~al\mbox{.}(2018)]%
        {bylinskii2018different}
\bibfield{author}{\bibinfo{person}{Zoya Bylinskii}, \bibinfo{person}{Tilke
  Judd}, \bibinfo{person}{Aude Oliva}, \bibinfo{person}{Antonio Torralba},
  {and} \bibinfo{person}{Fr{\'e}do Durand}.} \bibinfo{year}{2018}\natexlab{}.
\newblock \showarticletitle{What do different evaluation metrics tell us about
  saliency models?}
\newblock \bibinfo{journal}{\emph{IEEE transactions on pattern analysis and
  machine intelligence}} \bibinfo{volume}{41}, \bibinfo{number}{3}
  (\bibinfo{year}{2018}), \bibinfo{pages}{740--757}.
\newblock


\bibitem[Chang et~al\mbox{.}(2021)]%
        {chang2021mitigating}
\bibfield{author}{\bibinfo{person}{Jonathan Chang}, \bibinfo{person}{Masatoshi
  Uehara}, \bibinfo{person}{Dhruv Sreenivas}, \bibinfo{person}{Rahul Kidambi},
  {and} \bibinfo{person}{Wen Sun}.} \bibinfo{year}{2021}\natexlab{}.
\newblock \showarticletitle{Mitigating covariate shift in imitation learning
  via offline data with partial coverage}.
\newblock \bibinfo{journal}{\emph{Advances in Neural Information Processing
  Systems}}  \bibinfo{volume}{34} (\bibinfo{year}{2021}),
  \bibinfo{pages}{965--979}.
\newblock


\bibitem[Chen et~al\mbox{.}(2019)]%
        {chen2019gaze}
\bibfield{author}{\bibinfo{person}{Yuying Chen}, \bibinfo{person}{Congcong
  Liu}, \bibinfo{person}{Lei Tai}, \bibinfo{person}{Ming Liu}, {and}
  \bibinfo{person}{Bertram~E Shi}.} \bibinfo{year}{2019}\natexlab{}.
\newblock \showarticletitle{Gaze training by modulated dropout improves
  imitation learning}. In \bibinfo{booktitle}{\emph{2019 IEEE/RSJ International
  Conference on Intelligent Robots and Systems (IROS)}}. IEEE,
  \bibinfo{pages}{7756--7761}.
\newblock


\bibitem[Chib and Singh(2023)]%
        {chib2023recent}
\bibfield{author}{\bibinfo{person}{Pranav~Singh Chib} {and}
  \bibinfo{person}{Pravendra Singh}.} \bibinfo{year}{2023}\natexlab{}.
\newblock \showarticletitle{Recent advancements in end-to-end autonomous
  driving using deep learning: A survey}.
\newblock \bibinfo{journal}{\emph{IEEE Transactions on Intelligent Vehicles}}
  (\bibinfo{year}{2023}).
\newblock


\bibitem[Codevilla et~al\mbox{.}(2018)]%
        {codevilla2018end}
\bibfield{author}{\bibinfo{person}{Felipe Codevilla}, \bibinfo{person}{Matthias
  M{\"u}ller}, \bibinfo{person}{Antonio L{\'o}pez}, \bibinfo{person}{Vladlen
  Koltun}, {and} \bibinfo{person}{Alexey Dosovitskiy}.}
  \bibinfo{year}{2018}\natexlab{}.
\newblock \showarticletitle{End-to-end driving via conditional imitation
  learning}. In \bibinfo{booktitle}{\emph{2018 IEEE international conference on
  robotics and automation (ICRA)}}. IEEE, \bibinfo{pages}{4693--4700}.
\newblock


\bibitem[Cordts et~al\mbox{.}(2016)]%
        {cordts2016cityscapes}
\bibfield{author}{\bibinfo{person}{Marius Cordts}, \bibinfo{person}{Mohamed
  Omran}, \bibinfo{person}{Sebastian Ramos}, \bibinfo{person}{Timo Rehfeld},
  \bibinfo{person}{Markus Enzweiler}, \bibinfo{person}{Rodrigo Benenson},
  \bibinfo{person}{Uwe Franke}, \bibinfo{person}{Stefan Roth}, {and}
  \bibinfo{person}{Bernt Schiele}.} \bibinfo{year}{2016}\natexlab{}.
\newblock \showarticletitle{The cityscapes dataset for semantic urban scene
  understanding}. In \bibinfo{booktitle}{\emph{Proceedings of the IEEE
  conference on computer vision and pattern recognition}}.
  \bibinfo{pages}{3213--3223}.
\newblock


\bibitem[Dagdanov et~al\mbox{.}(2023)]%
        {dagdanov2023self}
\bibfield{author}{\bibinfo{person}{Resul Dagdanov}, \bibinfo{person}{Halil
  Durmus}, {and} \bibinfo{person}{Nazim~Kemal Ure}.}
  \bibinfo{year}{2023}\natexlab{}.
\newblock \showarticletitle{Self-Improving Safety Performance of Reinforcement
  Learning Based Driving with Black-Box Verification Algorithms}. In
  \bibinfo{booktitle}{\emph{2023 IEEE International Conference on Robotics and
  Automation (ICRA)}}. IEEE, \bibinfo{pages}{5631--5637}.
\newblock


\bibitem[Deng et~al\mbox{.}(2023)]%
        {deng2023driving}
\bibfield{author}{\bibinfo{person}{Tao Deng}, \bibinfo{person}{Lianfang Jiang},
  \bibinfo{person}{Yi Shi}, \bibinfo{person}{Jiang Wu},
  \bibinfo{person}{Zhangbi Wu}, \bibinfo{person}{Shun Yan},
  \bibinfo{person}{Xianshi Zhang}, {and} \bibinfo{person}{Hongmei Yan}.}
  \bibinfo{year}{2023}\natexlab{}.
\newblock \showarticletitle{Driving Visual Saliency Prediction of Dynamic Night
  Scenes via a Spatio-Temporal Dual-Encoder Network}.
\newblock \bibinfo{journal}{\emph{IEEE Transactions on Intelligent
  Transportation Systems}} (\bibinfo{year}{2023}).
\newblock


\bibitem[Deng et~al\mbox{.}(2019)]%
        {deng2019drivers}
\bibfield{author}{\bibinfo{person}{Tao Deng}, \bibinfo{person}{Hongmei Yan},
  \bibinfo{person}{Long Qin}, \bibinfo{person}{Thuyen Ngo}, {and}
  \bibinfo{person}{BS Manjunath}.} \bibinfo{year}{2019}\natexlab{}.
\newblock \showarticletitle{How do drivers allocate their potential attention?
  Driving fixation prediction via convolutional neural networks}.
\newblock \bibinfo{journal}{\emph{IEEE Transactions on Intelligent
  Transportation Systems}} \bibinfo{volume}{21}, \bibinfo{number}{5}
  (\bibinfo{year}{2019}), \bibinfo{pages}{2146--2154}.
\newblock


\bibitem[Deng et~al\mbox{.}(2020)]%
        {deng_CDNN}
\bibfield{author}{\bibinfo{person}{Tao Deng}, \bibinfo{person}{Hongmei Yan},
  \bibinfo{person}{Long Qin}, \bibinfo{person}{Thuyen Ngo}, {and}
  \bibinfo{person}{BS Manjunath}.} \bibinfo{year}{2020}\natexlab{}.
\newblock \showarticletitle{How Do Drivers Allocate Their Potential Attention?
  Driving Fixation Prediction via Convolutional Neural Networks}.
\newblock \bibinfo{journal}{\emph{IEEE Transactions on Intelligent
  Transportation Systems}} \bibinfo{volume}{21}, \bibinfo{number}{5}
  (\bibinfo{date}{May} \bibinfo{year}{2020}), \bibinfo{pages}{2146--2154}.
\newblock
\showISSN{1524-9050}
\urldef\tempurl%
\url{https://doi.org/10.1109/TITS.2019.2915540}
\showDOI{\tempurl}


\bibitem[Deng et~al\mbox{.}(2016)]%
        {deng2016does}
\bibfield{author}{\bibinfo{person}{Tao Deng}, \bibinfo{person}{Kaifu Yang},
  \bibinfo{person}{Yongjie Li}, {and} \bibinfo{person}{Hongmei Yan}.}
  \bibinfo{year}{2016}\natexlab{}.
\newblock \showarticletitle{Where does the driver look? Top-down-based saliency
  detection in a traffic driving environment}.
\newblock \bibinfo{journal}{\emph{IEEE Transactions on Intelligent
  Transportation Systems}} \bibinfo{volume}{17}, \bibinfo{number}{7}
  (\bibinfo{year}{2016}), \bibinfo{pages}{2051--2062}.
\newblock


\bibitem[Dosovitskiy et~al\mbox{.}(2017)]%
        {dosovitskiy2017carla}
\bibfield{author}{\bibinfo{person}{Alexey Dosovitskiy}, \bibinfo{person}{German
  Ros}, \bibinfo{person}{Felipe Codevilla}, \bibinfo{person}{Antonio Lopez},
  {and} \bibinfo{person}{Vladlen Koltun}.} \bibinfo{year}{2017}\natexlab{}.
\newblock \showarticletitle{CARLA: An open urban driving simulator}. In
  \bibinfo{booktitle}{\emph{Conference on robot learning}}. PMLR,
  \bibinfo{pages}{1--16}.
\newblock


\bibitem[Failing and Theeuwes(2018)]%
        {failing2018selection}
\bibfield{author}{\bibinfo{person}{Michel Failing} {and} \bibinfo{person}{Jan
  Theeuwes}.} \bibinfo{year}{2018}\natexlab{}.
\newblock \showarticletitle{Selection history: How reward modulates selectivity
  of visual attention}.
\newblock \bibinfo{journal}{\emph{Psychonomic bulletin \& review}}
  \bibinfo{volume}{25}, \bibinfo{number}{2} (\bibinfo{year}{2018}),
  \bibinfo{pages}{514--538}.
\newblock


\bibitem[Fang et~al\mbox{.}(2019)]%
        {fang2019dada}
\bibfield{author}{\bibinfo{person}{Jianwu Fang}, \bibinfo{person}{Dingxin Yan},
  \bibinfo{person}{Jiahuan Qiao}, \bibinfo{person}{Jianru Xue},
  \bibinfo{person}{He Wang}, {and} \bibinfo{person}{Sen Li}.}
  \bibinfo{year}{2019}\natexlab{}.
\newblock \showarticletitle{Dada-2000: Can driving accident be predicted by
  driver attentionƒ analyzed by a benchmark}. In
  \bibinfo{booktitle}{\emph{2019 IEEE Intelligent Transportation Systems
  Conference (ITSC)}}. IEEE, \bibinfo{pages}{4303--4309}.
\newblock


\bibitem[Fang et~al\mbox{.}(2021)]%
        {fang2021dada}
\bibfield{author}{\bibinfo{person}{Jianwu Fang}, \bibinfo{person}{Dingxin Yan},
  \bibinfo{person}{Jiahuan Qiao}, \bibinfo{person}{Jianru Xue}, {and}
  \bibinfo{person}{Hongkai Yu}.} \bibinfo{year}{2021}\natexlab{}.
\newblock \showarticletitle{DADA: Driver attention prediction in driving
  accident scenarios}.
\newblock \bibinfo{journal}{\emph{IEEE transactions on intelligent
  transportation systems}} \bibinfo{volume}{23}, \bibinfo{number}{6}
  (\bibinfo{year}{2021}), \bibinfo{pages}{4959--4971}.
\newblock


\bibitem[Filos et~al\mbox{.}(2020)]%
        {filos2020can}
\bibfield{author}{\bibinfo{person}{Angelos Filos}, \bibinfo{person}{Panagiotis
  Tigkas}, \bibinfo{person}{Rowan McAllister}, \bibinfo{person}{Nicholas
  Rhinehart}, \bibinfo{person}{Sergey Levine}, {and} \bibinfo{person}{Yarin
  Gal}.} \bibinfo{year}{2020}\natexlab{}.
\newblock \showarticletitle{Can autonomous vehicles identify, recover from, and
  adapt to distribution shifts?}. In \bibinfo{booktitle}{\emph{International
  Conference on Machine Learning}}. PMLR, \bibinfo{pages}{3145--3153}.
\newblock


\bibitem[Fujimoto et~al\mbox{.}(2018)]%
        {fujimoto2018addressing}
\bibfield{author}{\bibinfo{person}{Scott Fujimoto}, \bibinfo{person}{Herke
  Hoof}, {and} \bibinfo{person}{David Meger}.} \bibinfo{year}{2018}\natexlab{}.
\newblock \showarticletitle{Addressing function approximation error in
  actor-critic methods}. In \bibinfo{booktitle}{\emph{International conference
  on machine learning}}. PMLR, \bibinfo{pages}{1587--1596}.
\newblock


\bibitem[Gopinath et~al\mbox{.}(2021)]%
        {gopinath2021maad}
\bibfield{author}{\bibinfo{person}{Deepak Gopinath}, \bibinfo{person}{Guy
  Rosman}, \bibinfo{person}{Simon Stent}, \bibinfo{person}{Katsuya Terahata},
  \bibinfo{person}{Luke Fletcher}, \bibinfo{person}{Brenna Argall}, {and}
  \bibinfo{person}{John Leonard}.} \bibinfo{year}{2021}\natexlab{}.
\newblock \showarticletitle{Maad: A model and dataset for" attended awareness"
  in driving}. In \bibinfo{booktitle}{\emph{Proceedings of the IEEE/CVF
  International Conference on Computer Vision}}. \bibinfo{pages}{3426--3436}.
\newblock


\bibitem[Greydanus et~al\mbox{.}(2018)]%
        {greydanus2018visualizing}
\bibfield{author}{\bibinfo{person}{Samuel Greydanus}, \bibinfo{person}{Anurag
  Koul}, \bibinfo{person}{Jonathan Dodge}, {and} \bibinfo{person}{Alan Fern}.}
  \bibinfo{year}{2018}\natexlab{}.
\newblock \showarticletitle{Visualizing and understanding atari agents}. In
  \bibinfo{booktitle}{\emph{International conference on machine learning}}.
  PMLR, \bibinfo{pages}{1792--1801}.
\newblock


\bibitem[Grigorescu et~al\mbox{.}(2020)]%
        {RN40}
\bibfield{author}{\bibinfo{person}{Sorin Grigorescu}, \bibinfo{person}{Bogdan
  Trasnea}, \bibinfo{person}{Tiberiu Cocias}, {and} \bibinfo{person}{Gigel
  Macesanu}.} \bibinfo{year}{2020}\natexlab{}.
\newblock \showarticletitle{A survey of deep learning techniques for autonomous
  driving}.
\newblock \bibinfo{journal}{\emph{Journal of Field Robotics}}
  \bibinfo{volume}{37}, \bibinfo{number}{3} (\bibinfo{year}{2020}),
  \bibinfo{pages}{362--386}.
\newblock
\showISSN{1556-4959 1556-4967}
\urldef\tempurl%
\url{https://doi.org/10.1002/rob.21918}
\showDOI{\tempurl}


\bibitem[Guo et~al\mbox{.}(2021)]%
        {guo2021machine}
\bibfield{author}{\bibinfo{person}{Suna~Sihang Guo}, \bibinfo{person}{Ruohan
  Zhang}, \bibinfo{person}{Bo Liu}, \bibinfo{person}{Yifeng Zhu},
  \bibinfo{person}{Dana Ballard}, \bibinfo{person}{Mary Hayhoe}, {and}
  \bibinfo{person}{Peter Stone}.} \bibinfo{year}{2021}\natexlab{}.
\newblock \showarticletitle{Machine versus human attention in deep
  reinforcement learning tasks}.
\newblock \bibinfo{journal}{\emph{Advances in Neural Information Processing
  Systems}}  \bibinfo{volume}{34} (\bibinfo{year}{2021}),
  \bibinfo{pages}{25370--25385}.
\newblock


\bibitem[Hu and Clune(2024)]%
        {hu2024thought}
\bibfield{author}{\bibinfo{person}{Shengran Hu} {and} \bibinfo{person}{Jeff
  Clune}.} \bibinfo{year}{2024}\natexlab{}.
\newblock \showarticletitle{Thought cloning: Learning to think while acting by
  imitating human thinking}.
\newblock \bibinfo{journal}{\emph{Advances in Neural Information Processing
  Systems}}  \bibinfo{volume}{36} (\bibinfo{year}{2024}).
\newblock


\bibitem[Itaya et~al\mbox{.}(2021)]%
        {itaya2021visual}
\bibfield{author}{\bibinfo{person}{Hidenori Itaya}, \bibinfo{person}{Tsubasa
  Hirakawa}, \bibinfo{person}{Takayoshi Yamashita}, \bibinfo{person}{Hironobu
  Fujiyoshi}, {and} \bibinfo{person}{Komei Sugiura}.}
  \bibinfo{year}{2021}\natexlab{}.
\newblock \showarticletitle{Visual explanation using attention mechanism in
  actor-critic-based deep reinforcement learning}. In
  \bibinfo{booktitle}{\emph{2021 International Joint Conference On Neural
  Networks (IJCNN)}}. IEEE, \bibinfo{pages}{1--10}.
\newblock


\bibitem[Joo and Kim(2019)]%
        {joo2019visualization}
\bibfield{author}{\bibinfo{person}{Ho-Taek Joo} {and}
  \bibinfo{person}{Kyung-Joong Kim}.} \bibinfo{year}{2019}\natexlab{}.
\newblock \showarticletitle{Visualization of deep reinforcement learning using
  grad-CAM: how AI plays atari games?}. In \bibinfo{booktitle}{\emph{2019 IEEE
  Conference on Games (CoG)}}. IEEE, \bibinfo{pages}{1--2}.
\newblock


\bibitem[Kasahara et~al\mbox{.}(2022)]%
        {kasahara2022look}
\bibfield{author}{\bibinfo{person}{Isaac Kasahara}, \bibinfo{person}{Simon
  Stent}, {and} \bibinfo{person}{Hyun~Soo Park}.}
  \bibinfo{year}{2022}\natexlab{}.
\newblock \showarticletitle{Look both ways: Self-supervising driver gaze
  estimation and road scene saliency}. In \bibinfo{booktitle}{\emph{European
  Conference on Computer Vision}}. Springer, \bibinfo{pages}{126--142}.
\newblock


\bibitem[Kotseruba and Tsotsos(2021)]%
        {kotseruba2021behavioral}
\bibfield{author}{\bibinfo{person}{Iuliia Kotseruba} {and}
  \bibinfo{person}{John~K Tsotsos}.} \bibinfo{year}{2021}\natexlab{}.
\newblock \showarticletitle{Behavioral research and practical models of
  drivers' attention}.
\newblock \bibinfo{journal}{\emph{arXiv preprint arXiv:2104.05677}}
  (\bibinfo{year}{2021}).
\newblock


\bibitem[Kotseruba and Tsotsos(2022)]%
        {kotseruba2022attention}
\bibfield{author}{\bibinfo{person}{Iuliia Kotseruba} {and}
  \bibinfo{person}{John~K Tsotsos}.} \bibinfo{year}{2022}\natexlab{}.
\newblock \showarticletitle{Attention for vision-based assistive and automated
  driving: A review of algorithms and datasets}.
\newblock \bibinfo{journal}{\emph{IEEE transactions on intelligent
  transportation systems}} \bibinfo{volume}{23}, \bibinfo{number}{11}
  (\bibinfo{year}{2022}), \bibinfo{pages}{19907--19928}.
\newblock


\bibitem[Kumar et~al\mbox{.}(2019)]%
        {kumar2019stabilizing}
\bibfield{author}{\bibinfo{person}{Aviral Kumar}, \bibinfo{person}{Justin Fu},
  \bibinfo{person}{Matthew Soh}, \bibinfo{person}{George Tucker}, {and}
  \bibinfo{person}{Sergey Levine}.} \bibinfo{year}{2019}\natexlab{}.
\newblock \showarticletitle{Stabilizing off-policy q-learning via bootstrapping
  error reduction}.
\newblock \bibinfo{journal}{\emph{Advances in Neural Information Processing
  Systems}}  \bibinfo{volume}{32} (\bibinfo{year}{2019}).
\newblock


\bibitem[K{\"u}mmerer et~al\mbox{.}(2015)]%
        {kummerer2015information}
\bibfield{author}{\bibinfo{person}{Matthias K{\"u}mmerer},
  \bibinfo{person}{Thomas~SA Wallis}, {and} \bibinfo{person}{Matthias Bethge}.}
  \bibinfo{year}{2015}\natexlab{}.
\newblock \showarticletitle{Information-theoretic model comparison unifies
  saliency metrics}.
\newblock \bibinfo{journal}{\emph{Proceedings of the National Academy of
  Sciences}} \bibinfo{volume}{112}, \bibinfo{number}{52}
  (\bibinfo{year}{2015}), \bibinfo{pages}{16054--16059}.
\newblock


\bibitem[Kuutti et~al\mbox{.}(2020)]%
        {kuutti2020survey}
\bibfield{author}{\bibinfo{person}{Sampo Kuutti}, \bibinfo{person}{Richard
  Bowden}, \bibinfo{person}{Yaochu Jin}, \bibinfo{person}{Phil Barber}, {and}
  \bibinfo{person}{Saber Fallah}.} \bibinfo{year}{2020}\natexlab{}.
\newblock \showarticletitle{A survey of deep learning applications to
  autonomous vehicle control}.
\newblock \bibinfo{journal}{\emph{IEEE Transactions on Intelligent
  Transportation Systems}} \bibinfo{volume}{22}, \bibinfo{number}{2}
  (\bibinfo{year}{2020}), \bibinfo{pages}{712--733}.
\newblock


\bibitem[Lai et~al\mbox{.}(2020)]%
        {lai2020understanding}
\bibfield{author}{\bibinfo{person}{Qiuxia Lai}, \bibinfo{person}{Salman Khan},
  \bibinfo{person}{Yongwei Nie}, \bibinfo{person}{Hanqiu Sun},
  \bibinfo{person}{Jianbing Shen}, {and} \bibinfo{person}{Ling Shao}.}
  \bibinfo{year}{2020}\natexlab{}.
\newblock \showarticletitle{Understanding more about human and machine
  attention in deep neural networks}.
\newblock \bibinfo{journal}{\emph{IEEE Transactions on Multimedia}}
  \bibinfo{volume}{23} (\bibinfo{year}{2020}), \bibinfo{pages}{2086--2099}.
\newblock


\bibitem[Lajunen and Summala(1995)]%
        {lajunen1995driving}
\bibfield{author}{\bibinfo{person}{Timo Lajunen} {and} \bibinfo{person}{Heikki
  Summala}.} \bibinfo{year}{1995}\natexlab{}.
\newblock \showarticletitle{Driving experience, personality, and skill and
  safety-motive dimensions in drivers' self-assessments}.
\newblock \bibinfo{journal}{\emph{Personality and individual differences}}
  \bibinfo{volume}{19}, \bibinfo{number}{3} (\bibinfo{year}{1995}),
  \bibinfo{pages}{307--318}.
\newblock


\bibitem[Le~Meur and Baccino(2013)]%
        {le2013methods}
\bibfield{author}{\bibinfo{person}{Olivier Le~Meur} {and}
  \bibinfo{person}{Thierry Baccino}.} \bibinfo{year}{2013}\natexlab{}.
\newblock \showarticletitle{Methods for comparing scanpaths and saliency maps:
  strengths and weaknesses}.
\newblock \bibinfo{journal}{\emph{Behavior research methods}}
  \bibinfo{volume}{45}, \bibinfo{number}{1} (\bibinfo{year}{2013}),
  \bibinfo{pages}{251--266}.
\newblock


\bibitem[Li et~al\mbox{.}(2022)]%
        {li2022efficient}
\bibfield{author}{\bibinfo{person}{Quanyi Li}, \bibinfo{person}{Zhenghao Peng},
  {and} \bibinfo{person}{Bolei Zhou}.} \bibinfo{year}{2022}\natexlab{}.
\newblock \showarticletitle{Efficient learning of safe driving policy via
  human-ai copilot optimization}.
\newblock \bibinfo{journal}{\emph{arXiv preprint arXiv:2202.10341}}
  (\bibinfo{year}{2022}).
\newblock


\bibitem[Lim and Liu(2009)]%
        {lim2009modeling}
\bibfield{author}{\bibinfo{person}{Ji~Hyoun Lim} {and} \bibinfo{person}{Yili
  Liu}.} \bibinfo{year}{2009}\natexlab{}.
\newblock \showarticletitle{Modeling the influences of cyclic top-down and
  bottom-up processes for reinforcement learning in eye movements}.
\newblock \bibinfo{journal}{\emph{IEEE Transactions on Systems, Man, and
  Cybernetics-Part A: Systems and Humans}} \bibinfo{volume}{39},
  \bibinfo{number}{4} (\bibinfo{year}{2009}), \bibinfo{pages}{706--714}.
\newblock


\bibitem[Lindsay(2020)]%
        {lindsay2020attention}
\bibfield{author}{\bibinfo{person}{Grace~W Lindsay}.}
  \bibinfo{year}{2020}\natexlab{}.
\newblock \showarticletitle{Attention in psychology, neuroscience, and machine
  learning}.
\newblock \bibinfo{journal}{\emph{Frontiers in computational neuroscience}}
  \bibinfo{volume}{14} (\bibinfo{year}{2020}), \bibinfo{pages}{29}.
\newblock


\bibitem[Liu et~al\mbox{.}(2019)]%
        {inproceedingsgaze}
\bibfield{author}{\bibinfo{person}{Congcong Liu}, \bibinfo{person}{Yuying
  Chen}, \bibinfo{person}{Lei Tai}, \bibinfo{person}{Haoyang Ye},
  \bibinfo{person}{Ming Liu}, {and} \bibinfo{person}{Bert Shi}.}
  \bibinfo{year}{2019}\natexlab{}.
\newblock \showarticletitle{A Gaze Model Improves Autonomous Driving}.
\newblock
\urldef\tempurl%
\url{https://doi.org/10.1145/3314111.3319846}
\showDOI{\tempurl}


\bibitem[Luo et~al\mbox{.}(2023)]%
        {luo2023human}
\bibfield{author}{\bibinfo{person}{Biao Luo}, \bibinfo{person}{Zhengke Wu},
  \bibinfo{person}{Fei Zhou}, {and} \bibinfo{person}{Bing-Chuan Wang}.}
  \bibinfo{year}{2023}\natexlab{}.
\newblock \showarticletitle{Human-in-the-loop reinforcement learning in
  continuous-action space}.
\newblock \bibinfo{journal}{\emph{IEEE Transactions on Neural Networks and
  Learning Systems}} (\bibinfo{year}{2023}).
\newblock


\bibitem[Mascharka et~al\mbox{.}(2018)]%
        {mascharka2018transparency}
\bibfield{author}{\bibinfo{person}{David Mascharka}, \bibinfo{person}{Philip
  Tran}, \bibinfo{person}{Ryan Soklaski}, {and} \bibinfo{person}{Arjun
  Majumdar}.} \bibinfo{year}{2018}\natexlab{}.
\newblock \showarticletitle{Transparency by design: Closing the gap between
  performance and interpretability in visual reasoning}. In
  \bibinfo{booktitle}{\emph{Proceedings of the IEEE conference on computer
  vision and pattern recognition}}. \bibinfo{pages}{4942--4950}.
\newblock


\bibitem[Mhasawade et~al\mbox{.}(2024)]%
        {mhasawade2024understanding}
\bibfield{author}{\bibinfo{person}{Vishwali Mhasawade}, \bibinfo{person}{Salman
  Rahman}, \bibinfo{person}{Zoe Haskell-Craig}, {and} \bibinfo{person}{Rumi
  Chunara}.} \bibinfo{year}{2024}\natexlab{}.
\newblock \showarticletitle{Understanding Disparities in Post Hoc Machine
  Learning Explanation}.
\newblock \bibinfo{journal}{\emph{arXiv preprint arXiv:2401.14539}}
  (\bibinfo{year}{2024}).
\newblock


\bibitem[Mnih et~al\mbox{.}(2013)]%
        {mnih2013playing}
\bibfield{author}{\bibinfo{person}{Volodymyr Mnih}, \bibinfo{person}{Koray
  Kavukcuoglu}, \bibinfo{person}{David Silver}, \bibinfo{person}{Alex Graves},
  \bibinfo{person}{Ioannis Antonoglou}, \bibinfo{person}{Daan Wierstra}, {and}
  \bibinfo{person}{Martin Riedmiller}.} \bibinfo{year}{2013}\natexlab{}.
\newblock \showarticletitle{Playing atari with deep reinforcement learning}.
\newblock \bibinfo{journal}{\emph{arXiv preprint arXiv:1312.5602}}
  (\bibinfo{year}{2013}).
\newblock


\bibitem[Mott et~al\mbox{.}(2019)]%
        {mott2019towards}
\bibfield{author}{\bibinfo{person}{Alexander Mott}, \bibinfo{person}{Daniel
  Zoran}, \bibinfo{person}{Mike Chrzanowski}, \bibinfo{person}{Daan Wierstra},
  {and} \bibinfo{person}{Danilo Jimenez~Rezende}.}
  \bibinfo{year}{2019}\natexlab{}.
\newblock \showarticletitle{Towards interpretable reinforcement learning using
  attention augmented agents}.
\newblock \bibinfo{journal}{\emph{Advances in neural information processing
  systems}}  \bibinfo{volume}{32} (\bibinfo{year}{2019}).
\newblock


\bibitem[Naidu et~al\mbox{.}(2020)]%
        {DBLP:journals/corr/abs-2010-03023}
\bibfield{author}{\bibinfo{person}{Rakshit Naidu}, \bibinfo{person}{Ankita
  Ghosh}, \bibinfo{person}{Yash Maurya}, \bibinfo{person}{Shamanth R.~Nayak K},
  {and} \bibinfo{person}{Soumya~Snigdha Kundu}.}
  \bibinfo{year}{2020}\natexlab{}.
\newblock \showarticletitle{{IS-CAM:} Integrated Score-CAM for axiomatic-based
  explanations}.
\newblock \bibinfo{journal}{\emph{CoRR}}  \bibinfo{volume}{abs/2010.03023}
  (\bibinfo{year}{2020}).
\newblock
\showeprint[arXiv]{2010.03023}
\urldef\tempurl%
\url{https://arxiv.org/abs/2010.03023}
\showURL{%
\tempurl}


\bibitem[Nikulin et~al\mbox{.}(2019)]%
        {nikulin2019free}
\bibfield{author}{\bibinfo{person}{Dmitry Nikulin}, \bibinfo{person}{Anastasia
  Ianina}, \bibinfo{person}{Vladimir Aliev}, {and} \bibinfo{person}{Sergey
  Nikolenko}.} \bibinfo{year}{2019}\natexlab{}.
\newblock \showarticletitle{Free-lunch saliency via attention in atari agents}.
  In \bibinfo{booktitle}{\emph{2019 IEEE/CVF International Conference on
  Computer Vision Workshop (ICCVW)}}. IEEE, \bibinfo{pages}{4240--4249}.
\newblock


\bibitem[Ou et~al\mbox{.}(2023)]%
        {ou2023fuzzy}
\bibfield{author}{\bibinfo{person}{Liang Ou}, \bibinfo{person}{Yu-Chen Chang},
  \bibinfo{person}{Yu-Kai Wang}, {and} \bibinfo{person}{Chin-Teng Lin}.}
  \bibinfo{year}{2023}\natexlab{}.
\newblock \showarticletitle{Fuzzy Centered Explainable Network for
  Reinforcement Learning}.
\newblock \bibinfo{journal}{\emph{IEEE Transactions on Fuzzy Systems}}
  \bibinfo{volume}{32}, \bibinfo{number}{1} (\bibinfo{year}{2023}),
  \bibinfo{pages}{203--213}.
\newblock


\bibitem[Palazzi et~al\mbox{.}(2018)]%
        {palazzi2018predicting}
\bibfield{author}{\bibinfo{person}{Andrea Palazzi}, \bibinfo{person}{Davide
  Abati}, \bibinfo{person}{Francesco Solera}, \bibinfo{person}{Rita Cucchiara},
  {et~al\mbox{.}}} \bibinfo{year}{2018}\natexlab{}.
\newblock \showarticletitle{Predicting the Driver's Focus of Attention: the DR
  (eye) VE Project}.
\newblock \bibinfo{journal}{\emph{IEEE transactions on pattern analysis and
  machine intelligence}} \bibinfo{volume}{41}, \bibinfo{number}{7}
  (\bibinfo{year}{2018}), \bibinfo{pages}{1720--1733}.
\newblock


\bibitem[Peters et~al\mbox{.}(2005)]%
        {peters2005components}
\bibfield{author}{\bibinfo{person}{Robert~J Peters}, \bibinfo{person}{Asha
  Iyer}, \bibinfo{person}{Laurent Itti}, {and} \bibinfo{person}{Christof
  Koch}.} \bibinfo{year}{2005}\natexlab{}.
\newblock \showarticletitle{Components of bottom-up gaze allocation in natural
  images}.
\newblock \bibinfo{journal}{\emph{Vision research}} \bibinfo{volume}{45},
  \bibinfo{number}{18} (\bibinfo{year}{2005}), \bibinfo{pages}{2397--2416}.
\newblock


\bibitem[Poletti et~al\mbox{.}(2017)]%
        {poletti2017selective}
\bibfield{author}{\bibinfo{person}{Martina Poletti}, \bibinfo{person}{Michele
  Rucci}, {and} \bibinfo{person}{Marisa Carrasco}.}
  \bibinfo{year}{2017}\natexlab{}.
\newblock \showarticletitle{Selective attention within the foveola}.
\newblock \bibinfo{journal}{\emph{Nature neuroscience}} \bibinfo{volume}{20},
  \bibinfo{number}{10} (\bibinfo{year}{2017}), \bibinfo{pages}{1413--1417}.
\newblock


\bibitem[Pomerleau(1988)]%
        {pomerleau1988alvinn}
\bibfield{author}{\bibinfo{person}{Dean~A Pomerleau}.}
  \bibinfo{year}{1988}\natexlab{}.
\newblock \showarticletitle{Alvinn: An autonomous land vehicle in a neural
  network}.
\newblock \bibinfo{journal}{\emph{Advances in neural information processing
  systems}}  \bibinfo{volume}{1} (\bibinfo{year}{1988}).
\newblock


\bibitem[Raffin et~al\mbox{.}(2021)]%
        {stable-baselines3}
\bibfield{author}{\bibinfo{person}{Antonin Raffin}, \bibinfo{person}{Ashley
  Hill}, \bibinfo{person}{Adam Gleave}, \bibinfo{person}{Anssi Kanervisto},
  \bibinfo{person}{Maximilian Ernestus}, {and} \bibinfo{person}{Noah Dormann}.}
  \bibinfo{year}{2021}\natexlab{}.
\newblock \showarticletitle{Stable-Baselines3: Reliable Reinforcement Learning
  Implementations}.
\newblock \bibinfo{journal}{\emph{Journal of Machine Learning Research}}
  \bibinfo{volume}{22}, \bibinfo{number}{268} (\bibinfo{year}{2021}),
  \bibinfo{pages}{1--8}.
\newblock
\urldef\tempurl%
\url{http://jmlr.org/papers/v22/20-1364.html}
\showURL{%
\tempurl}


\bibitem[Ronneberger et~al\mbox{.}(2015)]%
        {ronneberger2015u}
\bibfield{author}{\bibinfo{person}{Olaf Ronneberger}, \bibinfo{person}{Philipp
  Fischer}, {and} \bibinfo{person}{Thomas Brox}.}
  \bibinfo{year}{2015}\natexlab{}.
\newblock \showarticletitle{U-net: Convolutional networks for biomedical image
  segmentation}. In \bibinfo{booktitle}{\emph{Medical Image Computing and
  Computer-Assisted Intervention--MICCAI 2015: 18th International Conference,
  Munich, Germany, October 5-9, 2015, Proceedings, Part III 18}}. Springer,
  \bibinfo{pages}{234--241}.
\newblock


\bibitem[Saran et~al\mbox{.}(2020)]%
        {saran2020efficiently}
\bibfield{author}{\bibinfo{person}{Akanksha Saran}, \bibinfo{person}{Ruohan
  Zhang}, \bibinfo{person}{Elaine~Schaertl Short}, {and} \bibinfo{person}{Scott
  Niekum}.} \bibinfo{year}{2020}\natexlab{}.
\newblock \showarticletitle{Efficiently guiding imitation learning agents with
  human gaze}.
\newblock \bibinfo{journal}{\emph{arXiv preprint arXiv:2002.12500}}
  (\bibinfo{year}{2020}).
\newblock


\bibitem[Schaul et~al\mbox{.}(2015)]%
        {schaul2015prioritized}
\bibfield{author}{\bibinfo{person}{Tom Schaul}, \bibinfo{person}{John Quan},
  \bibinfo{person}{Ioannis Antonoglou}, {and} \bibinfo{person}{David Silver}.}
  \bibinfo{year}{2015}\natexlab{}.
\newblock \showarticletitle{Prioritized experience replay}.
\newblock \bibinfo{journal}{\emph{arXiv preprint arXiv:1511.05952}}
  (\bibinfo{year}{2015}).
\newblock


\bibitem[Schulman et~al\mbox{.}(2017)]%
        {DBLP:journals/corr/SchulmanWDRK17}
\bibfield{author}{\bibinfo{person}{John Schulman}, \bibinfo{person}{Filip
  Wolski}, \bibinfo{person}{Prafulla Dhariwal}, \bibinfo{person}{Alec Radford},
  {and} \bibinfo{person}{Oleg Klimov}.} \bibinfo{year}{2017}\natexlab{}.
\newblock \showarticletitle{Proximal Policy Optimization Algorithms}.
\newblock \bibinfo{journal}{\emph{CoRR}}  \bibinfo{volume}{abs/1707.06347}
  (\bibinfo{year}{2017}).
\newblock
\showeprint[arXiv]{1707.06347}
\urldef\tempurl%
\url{http://arxiv.org/abs/1707.06347}
\showURL{%
\tempurl}


\bibitem[Selvaraju et~al\mbox{.}(2017)]%
        {selvaraju2017grad}
\bibfield{author}{\bibinfo{person}{Ramprasaath~R Selvaraju},
  \bibinfo{person}{Michael Cogswell}, \bibinfo{person}{Abhishek Das},
  \bibinfo{person}{Ramakrishna Vedantam}, \bibinfo{person}{Devi Parikh}, {and}
  \bibinfo{person}{Dhruv Batra}.} \bibinfo{year}{2017}\natexlab{}.
\newblock \showarticletitle{Grad-cam: Visual explanations from deep networks
  via gradient-based localization}. In \bibinfo{booktitle}{\emph{Proceedings of
  the IEEE international conference on computer vision}}.
  \bibinfo{pages}{618--626}.
\newblock


\bibitem[Shannon(1948)]%
        {shannon1948mathematical}
\bibfield{author}{\bibinfo{person}{Claude~Elwood Shannon}.}
  \bibinfo{year}{1948}\natexlab{}.
\newblock \showarticletitle{A mathematical theory of communication}.
\newblock \bibinfo{journal}{\emph{The Bell system technical journal}}
  \bibinfo{volume}{27}, \bibinfo{number}{3} (\bibinfo{year}{1948}),
  \bibinfo{pages}{379--423}.
\newblock


\bibitem[Shao et~al\mbox{.}(2023a)]%
        {shao2023safety}
\bibfield{author}{\bibinfo{person}{Hao Shao}, \bibinfo{person}{Letian Wang},
  \bibinfo{person}{Ruobing Chen}, \bibinfo{person}{Hongsheng Li}, {and}
  \bibinfo{person}{Yu Liu}.} \bibinfo{year}{2023}\natexlab{a}.
\newblock \showarticletitle{Safety-enhanced autonomous driving using
  interpretable sensor fusion transformer}. In
  \bibinfo{booktitle}{\emph{Conference on Robot Learning}}. PMLR,
  \bibinfo{pages}{726--737}.
\newblock


\bibitem[Shao et~al\mbox{.}(2023b)]%
        {shao2023reasonnet}
\bibfield{author}{\bibinfo{person}{Hao Shao}, \bibinfo{person}{Letian Wang},
  \bibinfo{person}{Ruobing Chen}, \bibinfo{person}{Steven~L Waslander},
  \bibinfo{person}{Hongsheng Li}, {and} \bibinfo{person}{Yu Liu}.}
  \bibinfo{year}{2023}\natexlab{b}.
\newblock \showarticletitle{ReasonNet: End-to-End Driving with Temporal and
  Global Reasoning}. In \bibinfo{booktitle}{\emph{Proceedings of the IEEE/CVF
  Conference on Computer Vision and Pattern Recognition}}.
  \bibinfo{pages}{13723--13733}.
\newblock


\bibitem[Shen et~al\mbox{.}(2022)]%
        {shen2022cocatt}
\bibfield{author}{\bibinfo{person}{Yuan Shen}, \bibinfo{person}{Niviru
  Wijayaratne}, \bibinfo{person}{Pranav Sriram}, \bibinfo{person}{Aamir Hasan},
  \bibinfo{person}{Peter Du}, {and} \bibinfo{person}{Katherine
  Driggs-Campbell}.} \bibinfo{year}{2022}\natexlab{}.
\newblock \showarticletitle{CoCAtt: A cognitive-conditioned driver attention
  dataset}. In \bibinfo{booktitle}{\emph{2022 IEEE 25th International
  Conference on Intelligent Transportation Systems (ITSC)}}. IEEE,
  \bibinfo{pages}{32--39}.
\newblock


\bibitem[Shiferaw et~al\mbox{.}(2019)]%
        {shiferaw2019gaze}
\bibfield{author}{\bibinfo{person}{Brook~A Shiferaw}, \bibinfo{person}{David~P
  Crewther}, {and} \bibinfo{person}{Luke~A Downey}.}
  \bibinfo{year}{2019}\natexlab{}.
\newblock \showarticletitle{Gaze entropy measures detect alcohol-induced driver
  impairment}.
\newblock \bibinfo{journal}{\emph{Drug and alcohol dependence}}
  \bibinfo{volume}{204} (\bibinfo{year}{2019}), \bibinfo{pages}{107519}.
\newblock


\bibitem[Shiferaw et~al\mbox{.}(2018)]%
        {shiferaw2018stationary}
\bibfield{author}{\bibinfo{person}{Brook~A Shiferaw}, \bibinfo{person}{Luke~A
  Downey}, \bibinfo{person}{Justine Westlake}, \bibinfo{person}{Bronwyn
  Stevens}, \bibinfo{person}{Shantha~MW Rajaratnam}, \bibinfo{person}{David~J
  Berlowitz}, \bibinfo{person}{Phillip Swann}, {and} \bibinfo{person}{Mark~E
  Howard}.} \bibinfo{year}{2018}\natexlab{}.
\newblock \showarticletitle{Stationary gaze entropy predicts lane departure
  events in sleep-deprived drivers}.
\newblock \bibinfo{journal}{\emph{Scientific reports}} \bibinfo{volume}{8},
  \bibinfo{number}{1} (\bibinfo{year}{2018}), \bibinfo{pages}{2220}.
\newblock


\bibitem[Silver et~al\mbox{.}(2017)]%
        {silver2017mastering}
\bibfield{author}{\bibinfo{person}{David Silver}, \bibinfo{person}{Julian
  Schrittwieser}, \bibinfo{person}{Karen Simonyan}, \bibinfo{person}{Ioannis
  Antonoglou}, \bibinfo{person}{Aja Huang}, \bibinfo{person}{Arthur Guez},
  \bibinfo{person}{Thomas Hubert}, \bibinfo{person}{Lucas Baker},
  \bibinfo{person}{Matthew Lai}, \bibinfo{person}{Adrian Bolton},
  {et~al\mbox{.}}} \bibinfo{year}{2017}\natexlab{}.
\newblock \showarticletitle{Mastering the game of go without human knowledge}.
\newblock \bibinfo{journal}{\emph{nature}} \bibinfo{volume}{550},
  \bibinfo{number}{7676} (\bibinfo{year}{2017}), \bibinfo{pages}{354--359}.
\newblock


\bibitem[Silvera et~al\mbox{.}(2022)]%
        {silvera2022dreyevr}
\bibfield{author}{\bibinfo{person}{Gustavo Silvera}, \bibinfo{person}{Abhijat
  Biswas}, {and} \bibinfo{person}{Henny Admoni}.}
  \bibinfo{year}{2022}\natexlab{}.
\newblock \showarticletitle{DReyeVR: Democratizing Virtual Reality Driving
  Simulation for Behavioural \& Interaction Research}. In
  \bibinfo{booktitle}{\emph{Proceedings of the 2022 ACM/IEEE International
  Conference on Human-Robot Interaction}}. \bibinfo{pages}{639--643}.
\newblock


\bibitem[Spering(2022)]%
        {spering2022eye}
\bibfield{author}{\bibinfo{person}{Miriam Spering}.}
  \bibinfo{year}{2022}\natexlab{}.
\newblock \showarticletitle{Eye movements as a window into decision-making}.
\newblock \bibinfo{journal}{\emph{Annual review of vision science}}
  \bibinfo{volume}{8} (\bibinfo{year}{2022}), \bibinfo{pages}{427--448}.
\newblock


\bibitem[Stanton et~al\mbox{.}(2019)]%
        {stanton2019models}
\bibfield{author}{\bibinfo{person}{Neville~A Stanton}, \bibinfo{person}{Paul~M
  Salmon}, \bibinfo{person}{Guy~H Walker}, {and} \bibinfo{person}{Maggie
  Stanton}.} \bibinfo{year}{2019}\natexlab{}.
\newblock \showarticletitle{Models and methods for collision analysis: A
  comparison study based on the Uber collision with a pedestrian}.
\newblock \bibinfo{journal}{\emph{Safety Science}}  \bibinfo{volume}{120}
  (\bibinfo{year}{2019}), \bibinfo{pages}{117--128}.
\newblock


\bibitem[Swain and Ballard(1991)]%
        {swain1991color}
\bibfield{author}{\bibinfo{person}{Michael~J Swain} {and}
  \bibinfo{person}{Dana~H Ballard}.} \bibinfo{year}{1991}\natexlab{}.
\newblock \showarticletitle{Color indexing}.
\newblock \bibinfo{journal}{\emph{International journal of computer vision}}
  \bibinfo{volume}{7}, \bibinfo{number}{1} (\bibinfo{year}{1991}),
  \bibinfo{pages}{11--32}.
\newblock


\bibitem[Taamneh et~al\mbox{.}(2017)]%
        {taamneh2017multimodal}
\bibfield{author}{\bibinfo{person}{Salah Taamneh}, \bibinfo{person}{Panagiotis
  Tsiamyrtzis}, \bibinfo{person}{Malcolm Dcosta}, \bibinfo{person}{Pradeep
  Buddharaju}, \bibinfo{person}{Ashik Khatri}, \bibinfo{person}{Michael
  Manser}, \bibinfo{person}{Thomas Ferris}, \bibinfo{person}{Robert
  Wunderlich}, {and} \bibinfo{person}{Ioannis Pavlidis}.}
  \bibinfo{year}{2017}\natexlab{}.
\newblock \showarticletitle{A multimodal dataset for various forms of
  distracted driving}.
\newblock \bibinfo{journal}{\emph{Scientific data}} \bibinfo{volume}{4},
  \bibinfo{number}{1} (\bibinfo{year}{2017}), \bibinfo{pages}{1--21}.
\newblock


\bibitem[Tang et~al\mbox{.}(2020)]%
        {tang2020neuroevolution}
\bibfield{author}{\bibinfo{person}{Yujin Tang}, \bibinfo{person}{Duong Nguyen},
  {and} \bibinfo{person}{David Ha}.} \bibinfo{year}{2020}\natexlab{}.
\newblock \showarticletitle{Neuroevolution of self-interpretable agents}. In
  \bibinfo{booktitle}{\emph{Proceedings of the 2020 Genetic and Evolutionary
  Computation Conference}}. \bibinfo{pages}{414--424}.
\newblock


\bibitem[Toromanoff et~al\mbox{.}(2020)]%
        {toromanoff2020end}
\bibfield{author}{\bibinfo{person}{Marin Toromanoff}, \bibinfo{person}{Emilie
  Wirbel}, {and} \bibinfo{person}{Fabien Moutarde}.}
  \bibinfo{year}{2020}\natexlab{}.
\newblock \showarticletitle{End-to-end model-free reinforcement learning for
  urban driving using implicit affordances}. In
  \bibinfo{booktitle}{\emph{Proceedings of the IEEE/CVF conference on computer
  vision and pattern recognition}}. \bibinfo{pages}{7153--7162}.
\newblock


\bibitem[Vaswani et~al\mbox{.}(2017)]%
        {vaswani2017attention}
\bibfield{author}{\bibinfo{person}{Ashish Vaswani}, \bibinfo{person}{Noam
  Shazeer}, \bibinfo{person}{Niki Parmar}, \bibinfo{person}{Jakob Uszkoreit},
  \bibinfo{person}{Llion Jones}, \bibinfo{person}{Aidan~N Gomez},
  \bibinfo{person}{{\L}ukasz Kaiser}, {and} \bibinfo{person}{Illia
  Polosukhin}.} \bibinfo{year}{2017}\natexlab{}.
\newblock \showarticletitle{Attention is all you need}.
\newblock \bibinfo{journal}{\emph{Advances in neural information processing
  systems}}  \bibinfo{volume}{30} (\bibinfo{year}{2017}).
\newblock


\bibitem[Wang et~al\mbox{.}(2020)]%
        {wang2020score}
\bibfield{author}{\bibinfo{person}{Haofan Wang}, \bibinfo{person}{Zifan Wang},
  \bibinfo{person}{Mengnan Du}, \bibinfo{person}{Fan Yang},
  \bibinfo{person}{Zijian Zhang}, \bibinfo{person}{Sirui Ding},
  \bibinfo{person}{Piotr Mardziel}, {and} \bibinfo{person}{Xia Hu}.}
  \bibinfo{year}{2020}\natexlab{}.
\newblock \showarticletitle{Score-CAM: Score-weighted visual explanations for
  convolutional neural networks}. In \bibinfo{booktitle}{\emph{Proceedings of
  the IEEE/CVF conference on computer vision and pattern recognition
  workshops}}. \bibinfo{pages}{24--25}.
\newblock


\bibitem[Wang et~al\mbox{.}(2022)]%
        {wang2022velocity}
\bibfield{author}{\bibinfo{person}{Zhe Wang}, \bibinfo{person}{Helai Huang},
  \bibinfo{person}{Jinjun Tang}, \bibinfo{person}{Xianwei Meng}, {and}
  \bibinfo{person}{Lipeng Hu}.} \bibinfo{year}{2022}\natexlab{}.
\newblock \showarticletitle{Velocity control in car-following behavior with
  autonomous vehicles using reinforcement learning}.
\newblock \bibinfo{journal}{\emph{Accident Analysis \& Prevention}}
  \bibinfo{volume}{174} (\bibinfo{year}{2022}), \bibinfo{pages}{106729}.
\newblock


\bibitem[Wen et~al\mbox{.}(2021)]%
        {wen2021keframe}
\bibfield{author}{\bibinfo{person}{Chuan Wen}, \bibinfo{person}{Jierui Lin},
  \bibinfo{person}{Jianing Qian}, \bibinfo{person}{Yang Gao}, {and}
  \bibinfo{person}{Dinesh Jayaraman}.} \bibinfo{year}{2021}\natexlab{}.
\newblock \showarticletitle{Keyframe-Focused Visual Imitation Learning}. In
  \bibinfo{booktitle}{\emph{Proceedings of the 38th International Conference on
  Machine Learning}} \emph{(\bibinfo{series}{Proceedings of Machine Learning
  Research}, Vol.~\bibinfo{volume}{139})}. \bibinfo{publisher}{PMLR},
  \bibinfo{pages}{11123--11133}.
\newblock


\bibitem[Wispinski et~al\mbox{.}(2020)]%
        {wispinski2020models}
\bibfield{author}{\bibinfo{person}{Nathan~J Wispinski},
  \bibinfo{person}{Jason~P Gallivan}, {and} \bibinfo{person}{Craig~S Chapman}.}
  \bibinfo{year}{2020}\natexlab{}.
\newblock \showarticletitle{Models, movements, and minds: bridging the gap
  between decision making and action}.
\newblock \bibinfo{journal}{\emph{Annals of the New York Academy of Sciences}}
  \bibinfo{volume}{1464}, \bibinfo{number}{1} (\bibinfo{year}{2020}),
  \bibinfo{pages}{30--51}.
\newblock


\bibitem[Wolfe(2010)]%
        {wolfe2010visual}
\bibfield{author}{\bibinfo{person}{Jeremy~M Wolfe}.}
  \bibinfo{year}{2010}\natexlab{}.
\newblock \showarticletitle{Visual search}.
\newblock \bibinfo{journal}{\emph{Current biology}} \bibinfo{volume}{20},
  \bibinfo{number}{8} (\bibinfo{year}{2010}), \bibinfo{pages}{R346--R349}.
\newblock


\bibitem[Wu et~al\mbox{.}(2023a)]%
        {wu2023toward}
\bibfield{author}{\bibinfo{person}{Jingda Wu}, \bibinfo{person}{Zhiyu Huang},
  \bibinfo{person}{Zhongxu Hu}, {and} \bibinfo{person}{Chen Lv}.}
  \bibinfo{year}{2023}\natexlab{a}.
\newblock \showarticletitle{Toward human-in-the-loop AI: Enhancing deep
  reinforcement learning via real-time human guidance for autonomous driving}.
\newblock \bibinfo{journal}{\emph{Engineering}}  \bibinfo{volume}{21}
  (\bibinfo{year}{2023}), \bibinfo{pages}{75--91}.
\newblock


\bibitem[Wu et~al\mbox{.}(2022)]%
        {wu2022prioritized}
\bibfield{author}{\bibinfo{person}{Jingda Wu}, \bibinfo{person}{Zhiyu Huang},
  \bibinfo{person}{Wenhui Huang}, {and} \bibinfo{person}{Chen Lv}.}
  \bibinfo{year}{2022}\natexlab{}.
\newblock \showarticletitle{Prioritized experience-based reinforcement learning
  with human guidance for autonomous driving}.
\newblock \bibinfo{journal}{\emph{IEEE Transactions on Neural Networks and
  Learning Systems}} (\bibinfo{year}{2022}).
\newblock


\bibitem[Wu et~al\mbox{.}(2023b)]%
        {wu2023human}
\bibfield{author}{\bibinfo{person}{Jingda Wu}, \bibinfo{person}{Yanxin Zhou},
  \bibinfo{person}{Haohan Yang}, \bibinfo{person}{Zhiyu Huang}, {and}
  \bibinfo{person}{Chen Lv}.} \bibinfo{year}{2023}\natexlab{b}.
\newblock \showarticletitle{Human-guided reinforcement learning with
  sim-to-real transfer for autonomous navigation}.
\newblock \bibinfo{journal}{\emph{IEEE Transactions on Pattern Analysis and
  Machine Intelligence}} (\bibinfo{year}{2023}).
\newblock


\bibitem[Xia et~al\mbox{.}(2019)]%
        {xia2019predicting}
\bibfield{author}{\bibinfo{person}{Ye Xia}, \bibinfo{person}{Danqing Zhang},
  \bibinfo{person}{Jinkyu Kim}, \bibinfo{person}{Ken Nakayama},
  \bibinfo{person}{Karl Zipser}, {and} \bibinfo{person}{David Whitney}.}
  \bibinfo{year}{2019}\natexlab{}.
\newblock \showarticletitle{Predicting driver attention in critical
  situations}. In \bibinfo{booktitle}{\emph{Computer Vision--ACCV 2018: 14th
  Asian Conference on Computer Vision, Perth, Australia, December 2--6, 2018,
  Revised Selected Papers, Part V 14}}. Springer, \bibinfo{pages}{658--674}.
\newblock


\bibitem[Xie et~al\mbox{.}(2021)]%
        {xie2021segformer}
\bibfield{author}{\bibinfo{person}{Enze Xie}, \bibinfo{person}{Wenhai Wang},
  \bibinfo{person}{Zhiding Yu}, \bibinfo{person}{Anima Anandkumar},
  \bibinfo{person}{Jose~M Alvarez}, {and} \bibinfo{person}{Ping Luo}.}
  \bibinfo{year}{2021}\natexlab{}.
\newblock \showarticletitle{SegFormer: Simple and efficient design for semantic
  segmentation with transformers}.
\newblock \bibinfo{journal}{\emph{Advances in neural information processing
  systems}}  \bibinfo{volume}{34} (\bibinfo{year}{2021}),
  \bibinfo{pages}{12077--12090}.
\newblock


\bibitem[Yu et~al\mbox{.}(2023)]%
        {yu2023offline}
\bibfield{author}{\bibinfo{person}{Lantao Yu}, \bibinfo{person}{Tianhe Yu},
  \bibinfo{person}{Jiaming Song}, \bibinfo{person}{Willie Neiswanger}, {and}
  \bibinfo{person}{Stefano Ermon}.} \bibinfo{year}{2023}\natexlab{}.
\newblock \showarticletitle{Offline imitation learning with suboptimal
  demonstrations via relaxed distribution matching}. In
  \bibinfo{booktitle}{\emph{Proceedings of the AAAI conference on artificial
  intelligence}}, Vol.~\bibinfo{volume}{37}. \bibinfo{pages}{11016--11024}.
\newblock


\bibitem[Zablocki et~al\mbox{.}(2022)]%
        {zablocki2022explainability}
\bibfield{author}{\bibinfo{person}{{\'E}loi Zablocki},
  \bibinfo{person}{H{\'e}di Ben-Younes}, \bibinfo{person}{Patrick P{\'e}rez},
  {and} \bibinfo{person}{Matthieu Cord}.} \bibinfo{year}{2022}\natexlab{}.
\newblock \showarticletitle{Explainability of deep vision-based autonomous
  driving systems: Review and challenges}.
\newblock \bibinfo{journal}{\emph{International Journal of Computer Vision}}
  \bibinfo{volume}{130}, \bibinfo{number}{10} (\bibinfo{year}{2022}),
  \bibinfo{pages}{2425--2452}.
\newblock


\bibitem[Zambaldi et~al\mbox{.}(2018)]%
        {zambaldi2018deep}
\bibfield{author}{\bibinfo{person}{Vinicius Zambaldi}, \bibinfo{person}{David
  Raposo}, \bibinfo{person}{Adam Santoro}, \bibinfo{person}{Victor Bapst},
  \bibinfo{person}{Yujia Li}, \bibinfo{person}{Igor Babuschkin},
  \bibinfo{person}{Karl Tuyls}, \bibinfo{person}{David Reichert},
  \bibinfo{person}{Timothy Lillicrap}, \bibinfo{person}{Edward Lockhart},
  {et~al\mbox{.}}} \bibinfo{year}{2018}\natexlab{}.
\newblock \showarticletitle{Deep reinforcement learning with relational
  inductive biases}. In \bibinfo{booktitle}{\emph{International conference on
  learning representations}}.
\newblock


\bibitem[Zhang et~al\mbox{.}(2018b)]%
        {zhang2018learning}
\bibfield{author}{\bibinfo{person}{Luxin Zhang}, \bibinfo{person}{Ruohan
  Zhang}, \bibinfo{person}{Zhuode Liu}, \bibinfo{person}{Mary Hayhoe}, {and}
  \bibinfo{person}{Dana Ballard}.} \bibinfo{year}{2018}\natexlab{b}.
\newblock \showarticletitle{Learning attention model from human for visuomotor
  tasks}. In \bibinfo{booktitle}{\emph{Proceedings of the AAAI Conference on
  Artificial Intelligence}}, Vol.~\bibinfo{volume}{32}.
\newblock


\bibitem[Zhang et~al\mbox{.}(2017)]%
        {zhang2017attention}
\bibfield{author}{\bibinfo{person}{Ruohan Zhang}, \bibinfo{person}{Zhuode Liu},
  \bibinfo{person}{Mary~M Hayhoe}, {and} \bibinfo{person}{Dana~H Ballard}.}
  \bibinfo{year}{2017}\natexlab{}.
\newblock \showarticletitle{Attention guided deep imitation learning}.
\newblock \bibinfo{journal}{\emph{Cognitive Computational Neuroscience (CCN)}}
  (\bibinfo{year}{2017}).
\newblock


\bibitem[Zhang et~al\mbox{.}(2018a)]%
        {zhang2018agil}
\bibfield{author}{\bibinfo{person}{Ruohan Zhang}, \bibinfo{person}{Zhuode Liu},
  \bibinfo{person}{Luxin Zhang}, \bibinfo{person}{Jake~A Whritner},
  \bibinfo{person}{Karl~S Muller}, \bibinfo{person}{Mary~M Hayhoe}, {and}
  \bibinfo{person}{Dana~H Ballard}.} \bibinfo{year}{2018}\natexlab{a}.
\newblock \showarticletitle{Agil: Learning attention from human for visuomotor
  tasks}. In \bibinfo{booktitle}{\emph{Proceedings of the european conference
  on computer vision (eccv)}}. \bibinfo{pages}{663--679}.
\newblock


\bibitem[Zhang et~al\mbox{.}(2020)]%
        {zhang2020atari}
\bibfield{author}{\bibinfo{person}{Ruohan Zhang}, \bibinfo{person}{Calen
  Walshe}, \bibinfo{person}{Zhuode Liu}, \bibinfo{person}{Lin Guan},
  \bibinfo{person}{Karl Muller}, \bibinfo{person}{Jake Whritner},
  \bibinfo{person}{Luxin Zhang}, \bibinfo{person}{Mary Hayhoe}, {and}
  \bibinfo{person}{Dana Ballard}.} \bibinfo{year}{2020}\natexlab{}.
\newblock \showarticletitle{Atari-head: Atari human eye-tracking and
  demonstration dataset}. In \bibinfo{booktitle}{\emph{Proceedings of the AAAI
  conference on artificial intelligence}}, Vol.~\bibinfo{volume}{34}.
  \bibinfo{pages}{6811--6820}.
\newblock


\bibitem[Zheng et~al\mbox{.}(2022)]%
        {zheng2022imitation}
\bibfield{author}{\bibinfo{person}{Boyuan Zheng}, \bibinfo{person}{Sunny
  Verma}, \bibinfo{person}{Jianlong Zhou}, \bibinfo{person}{Ivor~W Tsang},
  {and} \bibinfo{person}{Fang Chen}.} \bibinfo{year}{2022}\natexlab{}.
\newblock \showarticletitle{Imitation learning: Progress, taxonomies and
  challenges}.
\newblock \bibinfo{journal}{\emph{IEEE Transactions on Neural Networks and
  Learning Systems}} \bibinfo{number}{99} (\bibinfo{year}{2022}),
  \bibinfo{pages}{1--16}.
\newblock


\bibitem[Zhou et~al\mbox{.}(2016)]%
        {zhou2016cvpr}
\bibfield{author}{\bibinfo{person}{Bolei Zhou}, \bibinfo{person}{Aditya
  Khosla}, \bibinfo{person}{Agata Lapedriza}, \bibinfo{person}{Aude Oliva},
  {and} \bibinfo{person}{Antonio Torralba}.} \bibinfo{year}{2016}\natexlab{}.
\newblock \showarticletitle{Learning Deep Features for Discriminative
  Localization}. In \bibinfo{booktitle}{\emph{Computer Vision and Pattern
  Recognition}}.
\newblock


\end{thebibliography}
\end{document}